\documentclass[a4paper]{article}

\usepackage{amssymb,amsthm,amsmath}
\usepackage{xargs}                      

\usepackage[english]{babel}
\usepackage[utf8]{inputenc}
\usepackage[a4paper, top=30mm, bottom=30mm, right=22mm, left=22mm]{geometry}
\usepackage{graphicx}
\usepackage[font=small,labelfont=bf]{caption}
\usepackage{float}
\usepackage{calc}
\usepackage{isorot}
\usepackage[{width=0.89\textwidth},labelsep=period,singlelinecheck=false]{caption}

\usepackage{todonotes}
\newcommandx{\todomau}[2][1=]{\todo[linecolor=red,backgroundcolor=red!25,bordercolor=red,#1]{#2}}
\newcommandx{\todogianlu}[2][1=]{\todo[linecolor=red,backgroundcolor=green!25,bordercolor=red,#1]{#2}}

\usepackage[colorlinks,final,plainpages=false,pdfpagelabels]{hyperref}

\newcommand{\nine}{9$\times$9}
\newcommand{\seven}{7$\times$7}
\newcommand{\nineteen}{19$\times$19}
\newcommand{\dirfig}{./}
\newcommand{\dirtab}{\dirfig}

\title{SAI\\
  a Sensible Artificial Intelligence that plays with handicap\\
  and targets high scores in \nine\ Go }

\author{F.~Morandin, G.~Amato, M.~Fantozzi, R.~Gini, C.~Metta, M.~Parton}

\date{}

\begin{document}

\maketitle

\abstract{
\noindent We develop a new model that can be applied to any perfect information \mbox{two-player} zero-sum game to target a high score, and thus a perfect play. We integrate this model into the \mbox{Monte Carlo tree search-policy} iteration learning pipeline introduced by Google DeepMind with AlphaGo. Training this model on \nine\ Go produces a superhuman Go player, thus proving that it is stable and robust. We show that this model can be used to effectively play with both positional and score handicap, and to minimize suboptimal moves. We develop a family of agents that can target high scores against any opponent, and recover from very severe disadvantage against weak opponents.
To the best of our knowledge, these are the first effective achievements in this direction.
}

\tableofcontents

\section{Introduction}
The game of Go has been a landmark challenge for AI research since its
very beginning. It seems both very suited to AI, with the importance
of patterns and the need for deep exploration, and very tough to
actually solve, with its whole-board features and subtle
interdependencies of local situations. It is no surprise that DeepMind
first major effort and achievement was~\cite[AlphaGo]{AlphaGo}, an AI
that plays Go at superhuman level. It is nevertheless quite surprising
that the approach for this achievement works even better without human
knowledge~\cite[AlphaGo~Zero]{AlphaGoZero} and that it is universal enough to be
applied successfully to Chess and Shogi~\cite[AlphaZero]{AlphaZero}.


Despite the very specific application domain of Go, the AlphaGo family
achieved several quite general striking successes in Reinforcement
Learning (RL). They incorporated Monte Carlo tree search into a
generalized policy iteration scheme, for a better evaluation and
improvement step, in this way obtaining more realistic self-play
games and thus a more stable
learning process. They demonstrated that superhuman performance
in pure RL is possible, with minimal domain knowledge and no human
guidance. AlphaZero showed that this RL framework generalizes to
several perfect information two-players zero-sum games, suggesting
that these techniques are stable in the more general setting of
Multi-Agent RL -- a still very challenging framework suffering from
non-stationarity and theoretical infeasibility.

As a demonstration of the widespread influence that the AlphaGo family is
having on research, PubMed gives 177 published papers citing the first
AlphaGo paper from March 2016, and arXiv gives 191 preprint citing the
December 2017 AlphaZero preprint, as of May 24th.


In perfect information two-players zero-sum games, maximizing the
final score difference and the related abilities of playing with
positional or score handicap is still an open and important question.  For
instance, it would allow any player to be ranked according to the
score handicap needed to win $50\%$ of the times; it would allow human
players to learn an optimal game ending, and more generally optimal
sequences of moves in every situation; it would allow human players to
improve, by gradually decreasing the handicap.
AlphaGo is known to play suboptimal moves in the endgame, see for instance \cite[moves 210 and 214, page 252]{invisible}, and in general many games in \cite{invisible} not ending by resignation.
This phenomenon is rooted in the win/lose reward implemented in the Deep RL pipeline of AlphaGo. Score is unlikely to be a successful reward, because a single
point difference may change the winner, thus inducing instability in the training.
This has several downsides, see the detailed discussion in \cite[Introduction]{sai7x7}.

Efforts in the direction of score maximization have been made
in~\cite{DynamicKomi} and in \cite{CNNPrediction}. However, these
attempts don't use any of the modern DRL techniques, and thus their
accuracy is quite low. The only recent paper we are aware of is
\cite{MultipleKomi}, where a Deep Convolutional Neural Network is used to predict the final score and 41 different
winrates, corresponding to 41 different scores handicap
$\{-20,-19,\dots,0,\dots,19,20\}$. However, their
results have not been validated against human professional-level
players. Moreover, one single self-play training game is used to train
41 winrates, which is not a robust approach.
Finally, in~\cite{KataGo} the author introduces
a heavily modified implementation of AlphaGo Zero, which includes score estimation, among many
innovative features. The value to be
maximized is then a linear combination of winrate and expectation of a
nonlinear function of the score.
This could be a promising approach.

In this paper we present a new framework, called Sensible Artificial
Intelligence (SAI), that addresses the above-mentioned issues through a novel
modification of the AlphaGo framework: the winning probability is
modeled as a 2-parameters sigmoid, function of the targeted
score. Several self-play training games are branched, reinforcing robustness of this
model, see Section~\ref{s:branching}.  We first introduced this framework in a previous
work~\cite{sai7x7}, where we conducted a proof-of-concept in the toy
example of \seven\ Go. In this paper we could exploit the \nine\ setting to prove that 
the results promised in the \seven\ proof-of-concept have been indeed achieved, see the comment at the end of Section~\ref{s:sai7}.

The implementation of the SAI framework~\cite{SAIGitHub} has been realized as a fork
of Leela Zero~\cite{LeelaZero},  an open source clean room implementation of
AlphaGo Zero, by doubling the
value-head of the neural network, see Section~\ref{s:framework}.  We
performed two runs of the learning pipeline, from a random network to
very strong play on the \nine\ board, using two PCs with entry-level GPUs,
see Section~\ref{s:methods}.

We then apply massive simulations to show that SAI is able to play Go with both positional and score handicap, see
Section~\ref{s:handicap}, and can maximize the final score difference,
see Section~\ref{s:target_scores}. Moreover, SAI can leverage on its
strength to recover from very severe disadvantage when playing against
weaker opponents, see Section~\ref{s:dramatic}. All these features
have been validated against human players, see
Section~\ref{s:superhuman}: SAI won against a professional player and
against a professional-level player with substantial handicap,
therefore showing superhuman strength and additional ability to win
with handicap in games with humans. Finally, human validation shows that SAI minimizes suboptimal moves when compared with Leela Zero, see Section~\ref{s:suboptimal}.


A final remark: AlphaGo-like software doesn't play optimally when the
winrate is very high or very low, because in this case the winrate
cannot distinguish optimal from suboptimal moves (see for instance \cite[moves 210 and 214, page 252]{invisible}, and in general many games in \cite{invisible} not ending by resignation). Modeling the winrate
as a sigmoid makes it apparent that this is reminiscent of the ``vanishing gradient'' problem, which SAI agents solve by moving focus on a non-vanishing
area, see Section~\ref{s:agent_model}.

\section{The SAI framework}
\label{s:framework}

This section quickly recalls the principles, definitions and notation
introduced with more details in~\cite{sai7x7}.

\subsection{Modeling winrate as a sigmoid function of bonus points}
\label{s:sigmoid}

In the AlphaGo family the winning probability (or expected winrate)
of the current player depends on the game state $s$. In our
framework, we included an additional dependence on the number $x$ of
possible bonus points for the current player: in this way, trying to win
by $n$ points is equivalent to play trying to maximize the winrate in
$x=-n$. We modeled the winrate with a two-parameters sigmoid function,
as follows:
\begin{equation*}
\sigma_s(x)
:=\frac1{1+\exp(-\beta_s(\alpha_s+x))}
\end{equation*}

The number $\alpha_s$ is a shift parameter: since
$\sigma_s(-\alpha_s)=1/2$, it represents the expected difference of
points on the board from the perspective of the current player.  The
number $\beta_s$ is a scale parameter: the higher it is, the steeper
is the sigmoid, the higher the confidence that $\alpha_s$ is a good estimate 
of the difference in points, irrespective of the future moves.

\begin{figure}[ht]
  \caption{Examples of modeling the winrate for the current player as
    a function of the additional number of bonus points (that is,
    taking into account the $k_s=\pm k$ komi points). On the upper
    left the current player is winning, on the upper right the current
    player is losing. On the lower left the game is uncertain, on the
    lower right the game is decided and score is likely to be close to
    6.5 for the current player. }
	\label{f:examples}
	\begin{center}
		\includegraphics[width=.4\textwidth]{\dirfig/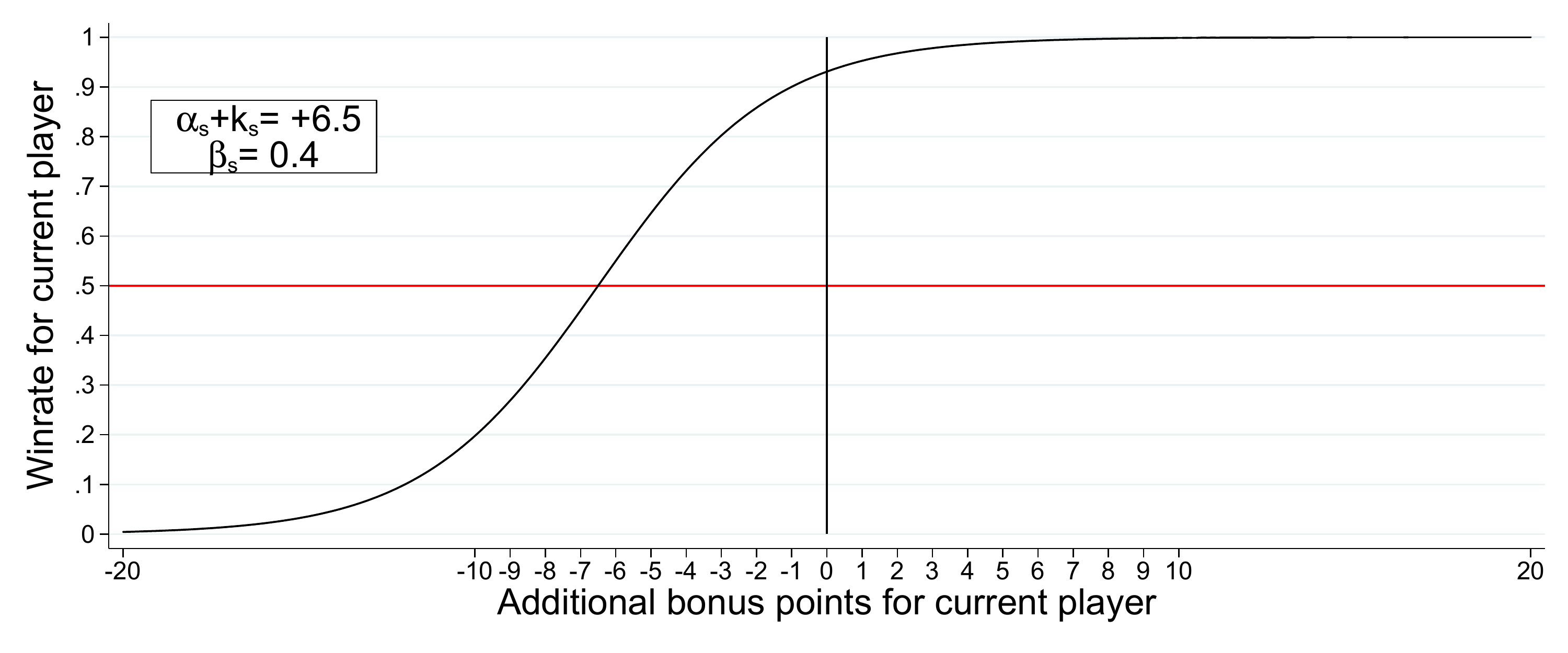}
		\includegraphics[width=.4\textwidth]{\dirfig/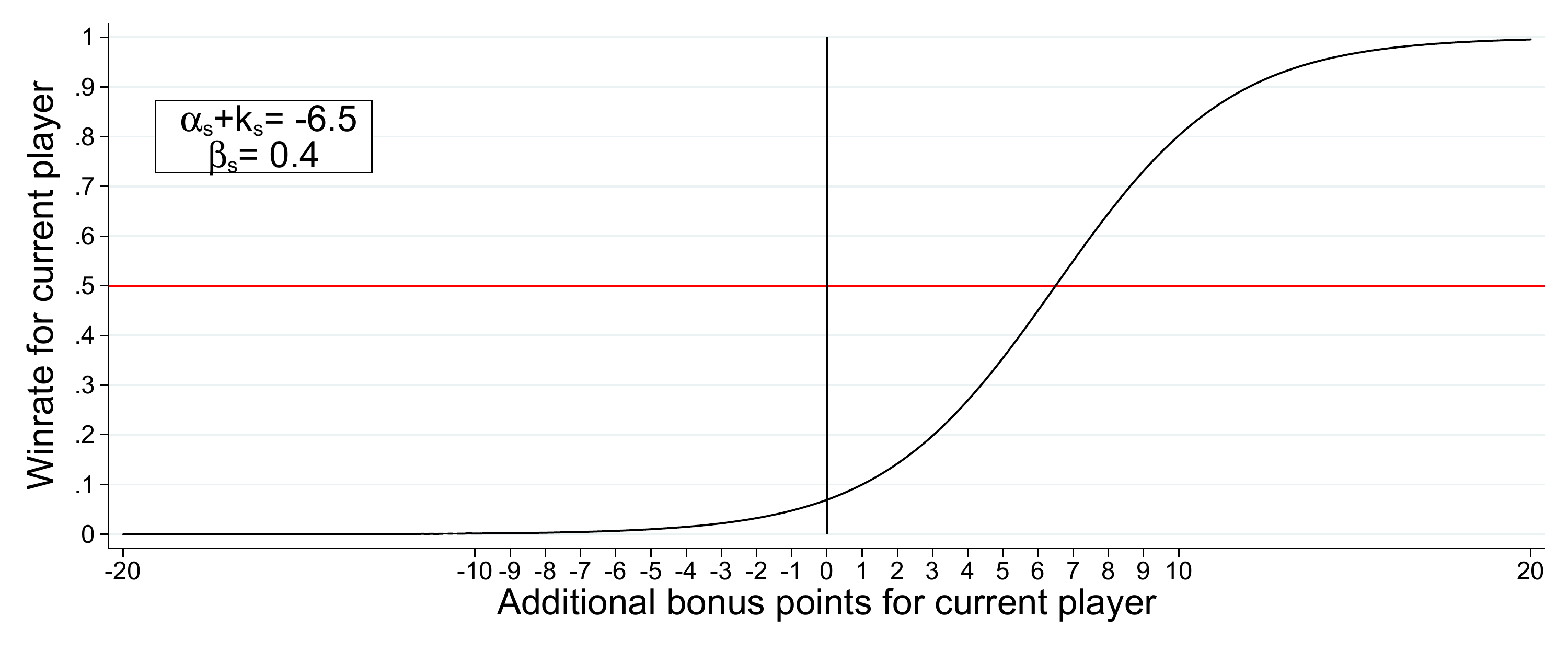}
		\includegraphics[width=.4\textwidth]{\dirfig/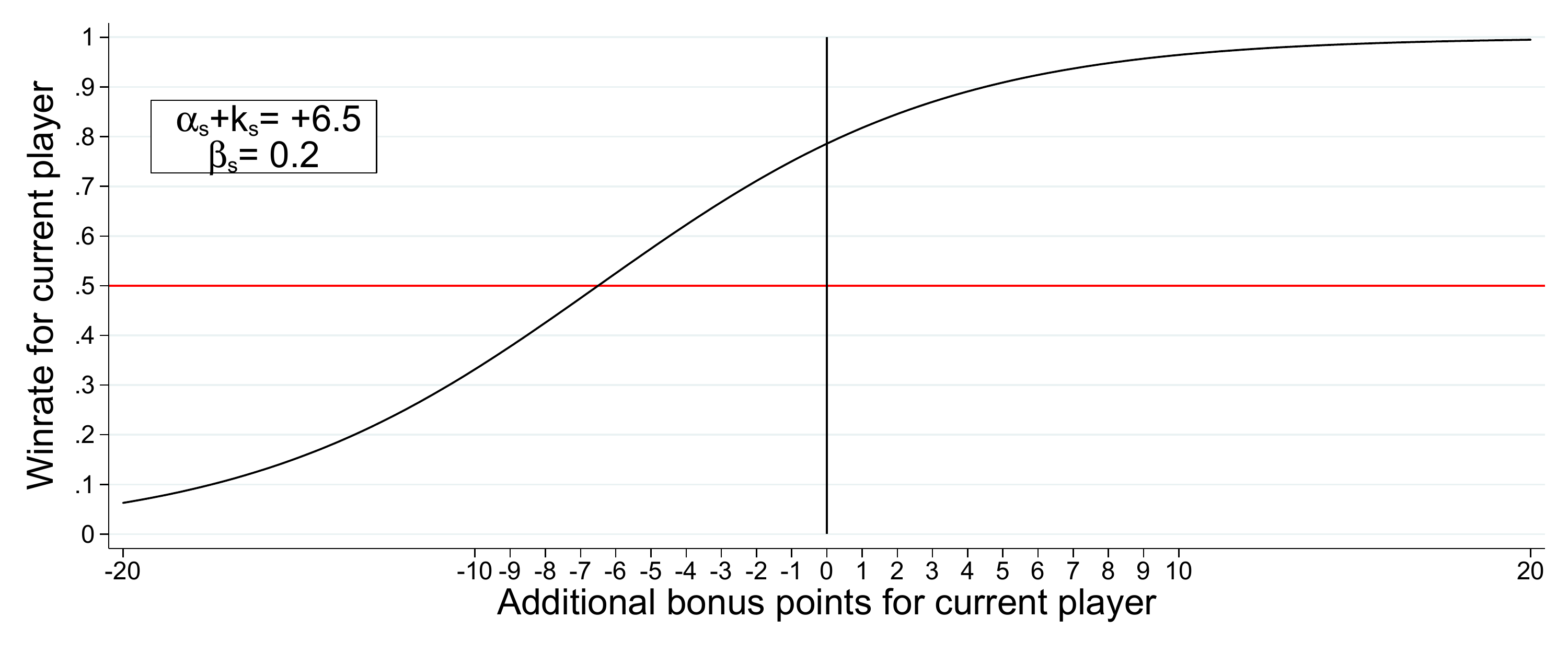}
		\includegraphics[width=.4\textwidth]{\dirfig/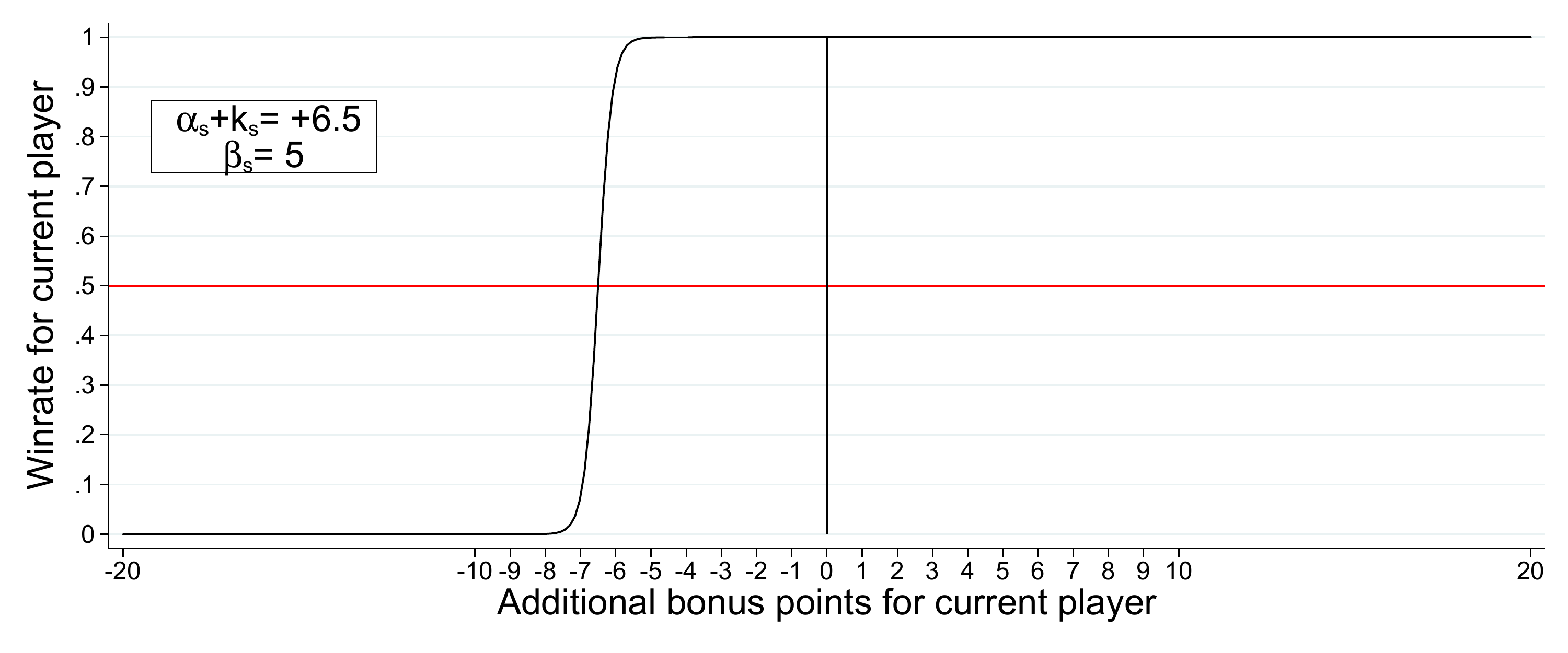}
	\end{center}
\end{figure}

Since Go is an intrinsically uneven game in favor of the first player,
which is traditionally black, a pre-specified amount $k$ of bonus
points, called \emph{komi}, is normally assigned to the white player
at the beginning of the game. For this reason, we incorporate the komi in the winrate function $\rho_s(x)$ as follows:
\begin{equation}\label{e:sigma_def}
 \rho_s(x)
  :=\sigma_s(x+k_s)=\frac1{1+\exp(-\beta_s(\alpha_s+k_s+x))}
\end{equation}
where $k_s$ denotes the \emph{signed komi}, relative to the current
player at $s$ (i.e.~$k_s=k$ if the current player is white and
$k_s=-k$ if the current player is black), and $x$ is the
\emph{additional} number of bonus points with respect to $k_s$.
The winning probability for the current player at position $s$ with komi $k_s$ is then $\rho_s(0)$, and the predicted score difference from the perspective of the current player is $\alpha_s+k_s$.
Figure~\ref{f:examples} illustrates some examples.

For instance, in a position $s$ of a game played with a komi of $k$, the winning probability of the current player with 3 additional bonus points is $\rho_s(3)$, and the winning probability of the current player with 4 additional malus points is $\rho_s(-4)$. 

\subsection{Duplicating the head of the neural network}
\label{s:model_nn}

AlphaGo, AlphaGo Zero, AlphaZero and Leela Zero all share the same core
structure, with neural networks
that for every
state $s$ provide:
\begin{itemize}
\item a probability distribution $p_s$ over the possible moves (the
  \emph{policy}), trained as to choose the most promising moves for
  searching the tree of subsequent positions;
\item a real number $v_s\in[0,1]$ (the \emph{value}), trained to
  estimate the winning probability for the current player.
\end{itemize}
In our framework, the neural network was modified to estimate, beyond
the usual policy $p_s$, the two parameters $\alpha_s$ and $\beta_s$ of
the sigmoid, instead of $v_s$. According to the choice introduced in Section~\ref{s:sigmoid}, the winning probability at the current komi $k_s$ is derived from $\alpha_s$ and $\beta_s$ as $\rho_s(0)$, where $\rho_s$ is given by \eqref{e:sigma_def}. Note, however, that $\alpha_s$ and $\beta_s$ are independent from $k_s$.

In the rest of the paper, we denote the values estimated by the neural network by $\hat\alpha_s$ and $\hat\beta_s$, and the corresponding estimated sigmoid given by inserting $\hat\alpha_s$ and $\hat\beta_s$ in \eqref{e:sigma_def}, by $\hat\rho_s$.  

\subsection{Branching from intermediate position}
\label{s:branching}


In order to train the two sigmoid parameters for each position, we relaxed the habit
of starting all training games from the initial
empty board position, and sometimes branched games at a certain state $s$,
changing the komi of the branch according to the current estimate of $\alpha_s$. In this way, we generated fragments of games with natural balanced situations
but a wide range of komi values. This reinforces robustness of the model.
Only a sample of all the possible positions were branched, nevertheless the network was able to generalize from this sample and obtain sharp estimates of $\alpha_s$, with high values of $\hat\beta_s$, for positions near the end of the game.

\subsection{Parametric family of value functions}
\label{s:agent_model}

In Leela Zero's Monte Carlo tree search, every playout that reaches a
state $s$ then chooses among the possible actions $a$ (identified
with children nodes) according to the policy $p_s(a)$ and to the
evaluation $Q(s,a)$ of the winrate. The latter is the \emph{average}
of the value $v(r)$ over the subtree of visited states $r$ rooted at
$a$. The choice is done according to AlphaGo Zero UCT formula,
see~\cite{AlphaGoZero,sai7x7}.
  
\begin{figure}[ht]
  \caption{If $s$ is a position and $a$ a possible move, the value
    $\nu_{\lambda,\mu}(a)$ is the average of the integrals of $\rho_r$
    between $x_\lambda$ and $x_\mu$, where $r$ are positions in a
    subtree rooted at $a$, and $x_\lambda$ and $x_\mu$ determined according 
    to \eqref{eq:lambdamu}.
  }
  \label{f:integral_agent}
  \begin{center}
    \includegraphics[width=.9\textwidth]{\dirfig/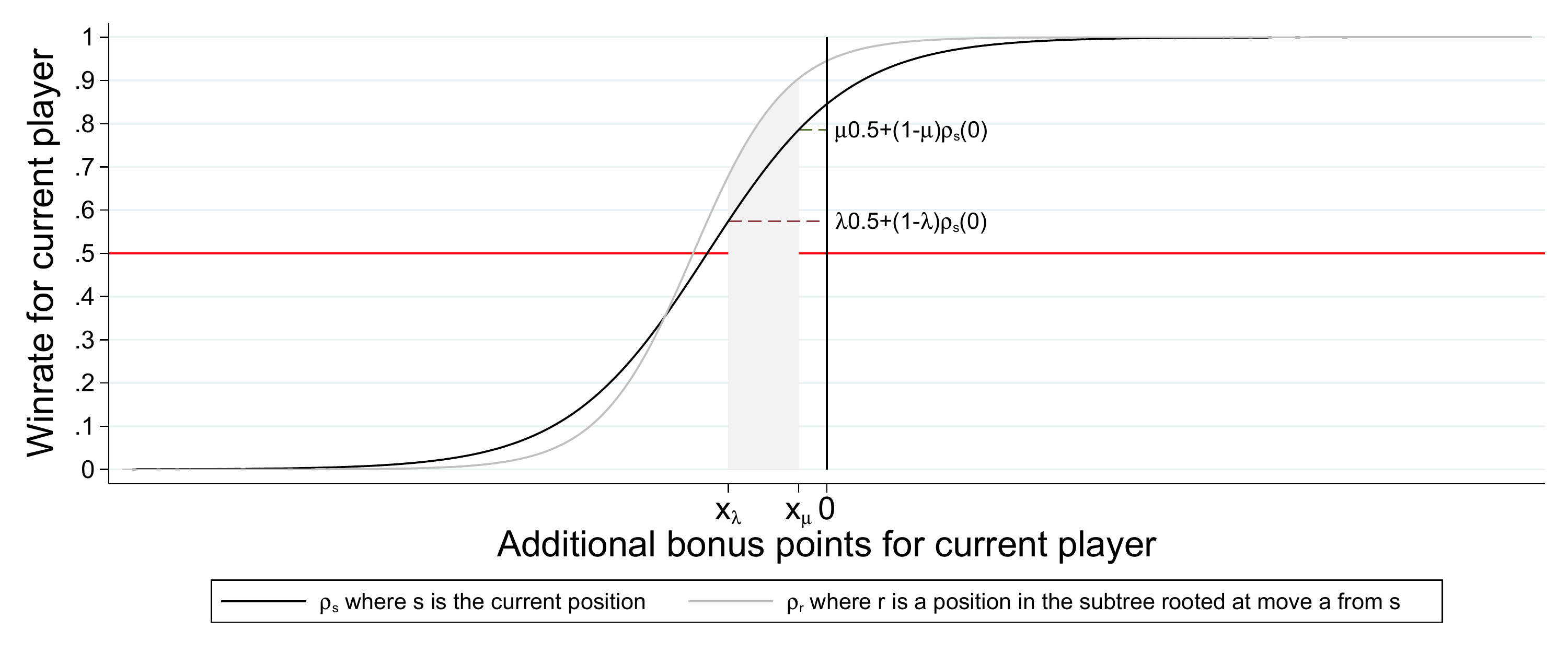}
  \end{center}
\end{figure}

In our framework, the value function equivalent to $v(r)$ is the value
$\rho_r(0)$ of the sigmoid~\eqref{e:sigma_def}. We designed an
additional parametric family of value functions
$\nu(r)=\nu_{\lambda,\mu}(r)$, $\lambda\geq\mu\in[0,1]$ computed as
\[
\nu_{\lambda,\mu}(r)
:=\begin{cases}
\frac1{x_\lambda-x_\mu}
\int_{x_\mu}^{x_\lambda}\rho_r(u)du
& \lambda>\mu\\
\rho_r(x_\lambda)
& \lambda=\mu
\end{cases}
\]
with $x_\lambda$ and $x_\mu$ pre-images via $\rho_s$ of convex
combinations between $\rho_s(0)$ and $0.5$, i.e.:
\begin{equation}
\label{eq:lambdamu}
\rho_{s}(x_\eta):=\eta0.5+ (1-\eta)\rho_{s}(0)
,\qquad \eta=\lambda,\mu
\end{equation}
Here, $x_\lambda$ and $x_\mu$ (and hence $\nu(r)$) are computed
according to the evaluation $\rho_{s}$ at some earlier node $s$, so
that the integral means $\nu(r)$ entering in the averages
$Q(s,a)$ in order to be compared, are all done on
the same interval. Remark that $x_0=0$ and $x_1=-(\alpha_s+k_s)$.
See Figure~\ref{f:integral_agent}.

The aim of these value functions is to better deal with situations in
which $\rho_r(0)=\nu_{0,0}(r)$ is close to 0 or to 1, when it is
difficult to distinguish good moves from moves that do not change the
winrate much, but are bad for the score. The integral average we
perform is useful to shift focus on a range of bonus points for which
the moves are easier to sort out.

We remark that for $\lambda>0$ and $\mu\in[0,\lambda]$,
$\nu_{\lambda,\mu}(r)$ under-estimates or over-estimates the winning
probability, according to whether the player's winrate is above or
below $0.5$. 
In the extreme scenario $\lambda=\mu=1$, the agent $\nu_{1,1}$ would always believe to be in a perfectly balanced situation. Thus, it would try to grab every single point, resulting in a greedy score-maximizing agent.

As we will show, when adopted, the parametric family $\nu_{\lambda,\mu}$ is
instrumental in pushing SAI towards higher scores.

\subsection{Results from the \seven\ proof-of-concept}
\label{s:sai7}

In our previous work we showed that our framework can successfully be
applied to \seven\ Go. In particular we remark four achievements,
obtained by the best networks of more than ten successful runs:
\begin{itemize}
\item The double value head estimated correctly both parameters in
  final and nearly final positions, with $\hat\alpha_s+k_s$ typically
  within $0.5$ points from the real score difference and $\hat\beta_s>5$.
  This is not trivial, since the target to which the value head is
  trained is not a pair $(\alpha,\beta)$, but the boolean game outcome
  at one single komi value, and in fact we couldn't get this result
  without the branching mechanism, which seems quite necessary to this
  end. This paper shows that this achievement was not due to the relative small number of 
  positions on \seven\ Go, see end of Section~\ref{s:scoring} and Figure~\ref{f:scoring}.
\item The sequences of moves chosen by the policy at the very
  beginning of each game correspond to what is believed to be the
  \emph{perfect play} (there are tentative solutions of \seven\ Go by
  strong professional players, see~\cite{Davies7x7,LiWei7x7}).
  Moreover the estimated parameters for the initial position were
  $\hat\alpha_\varnothing\approx9$ and $\hat\beta_\varnothing>2$, suggesting
  that the nets properly identified the fair komi of 9 points, and
  could win with 95\% confidence if given just 0.5 bonus points from
  it. We believe that the play at 250 visits was perfect or very near
  to perfect.
\item The learning pipeline was efficient and fast, with high
  reproducibility between runs. In particular 500 millions nodes
  (equivalent to about 7 millions moves, or 250,000 games) were
  sufficient to go from random networks to almost perfect play.
\item The overall complexity of the games was high, with a clear
  understanding of subtle, non-local game properties, such as
  \emph{seki} and \emph{ko} (at least at 250 visits), and this was
  achieved by nets with just 3 residual convolutional blocks. The
  performance did not improve by raising this value.
\end{itemize}

Finally we were able to test the use of the parameters of the value
function and we proved that in a winning position with score not yet
set, larger values of $\lambda$ increase the likelihood that the best
move is chosen.

However, \seven\ Go is never played at professional level, nor other
artificial intelligences exist that we could compare our nets to.
Moreover, \seven\ games are too simplistic and score differences are too
limited to experimenting properly with different agents. Finally, the
fact that almost perfect play can actually be learnt on \seven\ Go,
made the possibility of generalization to larger sizes at least
dubious.

In this work we exploited the setting of \nine\ Go. This allowed us to
challenge our assumptions and prove that our framework is effective.

\section{Methods of this paper}
\label{s:methods}

\subsection{Training a \nine\ SAI}

We performed two runs of \nine\ SAI to verify that the learning
pipeline would also work on this much more complex framework.

The process we implemented is similar to what was done in Leela
Zero~\cite{LeelaZero}, with a sequence of \emph{generations}, each one
with a single network doing self-play games, beginning with a random
net.
Differently from Leela Zero, in our setting there is no \emph{gating},
meaning that after a fixed number of games the generation ends and a
newly trained net is automatically promoted, without testing that it
wins against the previous one. The number of self-play games per generation is important for
efficiency, and after some initial experiments with larger values, we
confirmed that a number as small as 2,000 is
enough to get a fast and stable learning (with the possible exception 
of the first 3-4 generations).


Each training was performed on the self-play games data of a variable
number of generations, ranging from 4 to 20, inversely proportional to
the speed of the changes in the nets from generation to generation, so
that in the training buffer there would not be contradictory
information.

In each generation, the proportion of complete games to branches was
2:1. Complete games always spanned several komi values, chosen in
$\frac12\mathbb Z$ with a distribution obtained by interpreting the
sigmoid $\hat\rho_\varnothing$ (of the current net, for the empty board)
as a cumulative distribution function.  Branches were originated from
random positions $s$ (each position in a game or branch had the same
probability $p=0.02$ of originating a new branch) and new komi set
equal to $\pm\hat\alpha_s$ (rounded to half an integer) with the sign
appropriate for the color of the current player, so that the starting
position $s$ with the new komi would be estimated as fair for the two
players.

The training hyperparameters changed several times during the two
runs, with typical training rate 0.0001, batch size 512 and 4,000--10,000
training steps per generation. In the second run we experimented with
a kind of ``weak gating'', in the sense that for every generation
during the training we exported 10 networks, at regular intervals of
steps, and then match every one against the previous network, finally
promoting the one with the best performance. It is unclear if this
choice improves learning, but it seems to reduce strength oscillations.

\subsection{Network structure}

The structure of the neural networks was always the same during the runs, though with
different sizes. Its main part is the same as Leela Zero. The differences from Leela Zero are: 
the value head, that in SAI is a double head; the layer root of the value head, which is bigger; 
and some domain knowledge, added to increase the network awareness of non-local properties of the position. 
In the following we give the details of SAI network structure.

The input is formed by 17 bitplanes of size \nine: one is constant,
with 1 in all intersections (useful for the network to be aware of
borders, thanks to the zero-padding of subsequent convolutions). The
remaining planes hold 4 different features for the last 4 positions in
the game: current player stones, opponent stones, illegal moves, last
liberties of groups. The latter two are peculiar Go features that were
not originally contemplated in Leela Zero, but their introduction can
be very beneficial (see also~\cite{KataGo}). We remark that there is no
need to code the current player color, as the evaluations should be
valid for all komi values: in fact we confirmed in~\cite{sai7x7} that
this choice enhances somewhat the performances of a \seven\ SAI run.

The first layer is a $3\times3$ convolutional layer with $k$ filters,
followed by batch normalization and ReLU. Then there is a tower of $n$
identical blocks, each one a $3\times3$ residual convolutional layer
with $k$ filters followed by batch normalization and ReLU.

On top of that there are the two heads. The policy head is composed by
a $1\times1$ convolutional layer with 2 filters, batch normalization
and ReLU, followed by a dense layer with 82 outputs, one per move,
and then softmax.

The value head starts with a $1\times1$ convolutional layer with 3 or
2 filters (first and second run respectively), batch normalization and
ReLU, on top of which there are two almost identical sub-heads, for
$\alpha$ and $\beta$. Both sub-heads are composed by two dense
layers. The first layer has 384 or 256 outputs (for $\alpha$ or
$\beta$ sub-heads), and is followed by ReLU.  The second layer has
just 1 output. The $\beta$ sub-head is concluded by computing the
exponential of the last output.

The loss function is the sum of three terms: an $l^2$ regularization
term, the cross-entropy loss between the visits proportion and the
network estimate of the policy, and the mean-squared error loss
between the game result and the winrate estimate $\hat\rho_s(0)$.

\subsection{Scaling up complexity}

Since with limited computational resources the training of \nine\ SAI
is very long, we decided to start the process with simplified
settings, scaling up afterwards as the performance stalled.

This approach was introduced with success in Leela Zero, by increasing
progressively the network size, from $n=5$ blocks and $k=64$ filters
to $n=40$, $k=256$.

In \seven\ SAI it was observed that one could also progressively
increase the number $v$ of visits, as very small values are more
efficient at the beginning, while very large values may be needed for
getting to optimal play in the end. In particular it was noticed that
with low visits the progress could become stuck with a wrong sequence
of initial moves.

In the first run we started with $n=4$, $k=128$ and $v=100$ and
progressively increased visits to a maximum value of $v=850$. Then we
started increasing the network size to a maximum of $n=8$, $k=160$. In
total there were 690 generations, equivalent to about 1.5 millions
games, 70 million moves and 25 billions nodes.

In the second run we tried to keep the network structure large and
fixed at $k=256$, $n=12$ and scaled only the visits, starting from
a lower value of $v=25$ and going up to $v=400$ in 300 generations.

\subsection{Special issues}

The learning pipeline we implemented differs from Leela Zero in
several other choices and minor optimizations.

\begin{itemize}
\item We had the choice to set the agent parameters $\lambda$ and
  $\mu$ during self-plays. Though the effect is not completely
  understood, it was apparent that too high a value of these
  parameters would compromise performance, so we mostly set $\mu=0$
  and $\lambda=0$ or $\lambda=0.1$.
\item The initial 25 moves of each game (starting the count from the
  empty board position, in case of a branched game) were chosen
  randomly, according to the number of visits, with temperature 1
  (i.e.~proportionally) or less than 1. If a randomly chosen move was
  considered a ``blunder'' (we experimented with several definitions)
  the positions before the last blunder in the game were not recorded
  for training (since the epilogue of the game could be wrong for
  those positions).
\item It is believed that the general Leela Zero framework has a
  tendency, for the policy of the first moves, to converge to some
  fixed sequence that may be quasi-optimal. To avoid this we increased
  the depth of search in the initial position, by exploiting
  symmetries and keeping the visits both high (without early stop) and
  diversified (in the second run we experimented with a 75\% maximum
  visits for any move). Moreover we used a temperature higher than 1
  for evaluating the policy.
\end{itemize}

\subsection{Elo evaluation}

Every run of a project like Leela Zero or SAI yields hundreds of
neural networks, of substantially growing strength, but also with
oscillations and ``rock-paper-scissors'' triples. It is then quite
challenging to give them absolute numerical scores to measure their
performance.

The standard accepted metrics for human players is the Elo rating~\cite[Section~8.4]{EloRating}, defined up to an additive constant in
such a way that the difference of the rating of two players is
proportional to the \mbox{log-odds-ratio} of the first one winning:
\[
E_1-E_2:=400\log_{10}\left(\frac{P(\text{player 1 wins})}{P(\text{player 2 wins})}\right)
\]
Unfortunately it is not clear if the Elo model is rich enough to
account for all the results among networks. In particular it is
obviously not able to deal with rock-paper-scissors situations.

Nonetheless it is always possible to estimate an Elo rating difference
between two networks by making them play against each other. When this
is limited to matches between newly promoted networks and their
predecessor, as in Leela Zero, the resulting estimates are believed to
give an Elo rating inflation, in particular in combination with
gating.

Previously on \seven\ SAI, we adopted a multivariate evaluation by
matching networks against a fixed panel, see~\cite{sai7x7}, but
though highly informative, this approach lacks the universality and
reproducibility that are expected when dealing with the more
widespread setting of \nine\ Go.

Hence for \nine\ SAI we reverted to the Elo rating, but to get better
global estimates of the networks strengths, following~\cite{CloudyGo},
we confronted every network against several others, of comparable
ability, obtaining a graph of pairings with about 1,000 nodes and
13,000 edges. More specifically, every promoted network was compared
with those of the same run with generation differences
$\pm1,\pm2,\pm3,\pm6,\pm8,\pm12$, moreover one every four networks was
compared with 16 networks of a panel (slowly-changing within both
runs) and one every three networks was compared with the corresponding
generation of the other run.

The matches parameters were komi 7.5 (komi 7 in the second run),
$\lambda=\mu=0$, 225 visits, 30 or more games.

To estimate the Elo ratings of the networks, we implemented the
maximum likelihood estimator, in a way similar to the classic Bayesian
Elo Rating~\cite{BayesElo}, but with a specialization for dealing with
draws in the game of Go. Let $N$ be the set of players,
$s=(s_j)_{j\in N}$ a vector of ratings, let $M$ be the set of
matches. Define the log-likelihood as
\[
l(s)=\sum_{i\in M}\Bigl[a_i\log p_i(s) + b_i\log q_i(s) + c_i\log r_i(s)\Bigr]
\]
where $a_i,b_i,c_i$ are the number of wins, draws and losses in game
$i$ for the first player $f_i\in N$ against the second player
$g_i\in N$, and $p_i,q_i,r_i$ are estimates of the probabilities of
winning, tying and losing, obtained by maximizing the term
\[
a_i\log p_i + b_i\log q_i + c_i\log r_i
\]
under the constraints
\[
  \begin{cases}
  p_i+q_i+r_i
  =1 \\[1ex]
    400\log_{10}\left(\dfrac{1\cdot p_i+\frac12\cdot q_i+0\cdot r_i}
    {0\cdot p_i+\frac12\cdot q_i+1\cdot r_i}\right)
  =s_{f_i}-s_{g_i}
\end{cases}
\]
Then, by maximizing $l(s)$ as a function of $s$, one finds an MLE
$\hat s$ for the Elo ratings, that is a robust generalization
of~\cite{BayesElo} particularly good for the large number of match
games that can be generated by computer programs.

\subsection{Training outcome}\label{s:training_outcome}

While the real strength of SAI is its ability to deal with different
komi values, with handicap, and to target high score margins -- these
are discussed in the next section -- it is also important, during each
run, to check the bare playing strength progress.

\begin{figure}[ht]
  \caption{Estimate of the Elo rating as function of the generation
    for the two runs of \nine\ SAI. The nets $S1$, $S2$, $S3$, $W1$, $W2$, $W3$ are described in Section~\ref{s:experimental_setting}.}
	\label{f:elo}
	\begin{center}
		\includegraphics[width=.9\textwidth]{\dirfig/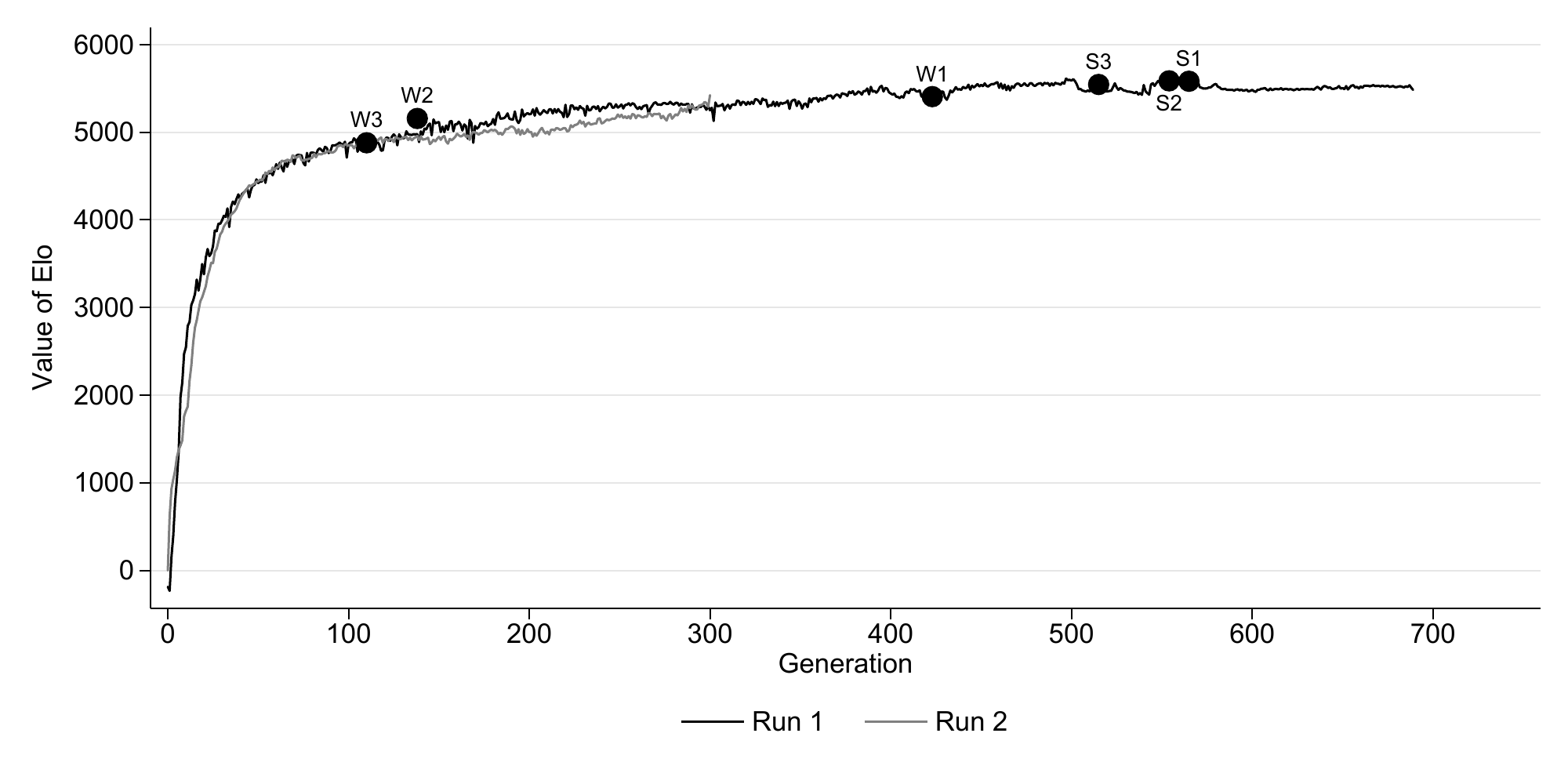}
	\end{center}
\end{figure}

Figure~\ref{f:elo}
shows the Elo rating of the networks of both runs, anchored to 0 for
the random network.
It is apparent that the growth is very fast at the beginning, ranging
from random play to a good amateur-level playing strength, and that
Elo rating is less able to express the subtle differences in the
playing style and game awareness of more mature networks, where
rock-paper-scissors situations are more frequent, and where, for
example, an exotic choice of the first moves by a network may balance
a lesser understanding of subsequent game situations.

The first run seems to oscillate rather widely, with occasional large
drops in performance that anyway disappear almost immediately. This
may be due to the complete absence of gating. Long-time oscillations
after generation 400 seem mainly due to changes in the first moves
preferences. The complexity of the games seemed to increase steadily
during both runs.

The second run was slightly more problematic, as some progress was
lost around generation 200, possibly because we upgraded the program
during the run and had to adapt some parameters. The local
oscillations were smaller and the progress slightly slower in the
first generations. The very last networks performance seems inflated,
possibly because recent networks are mainly matched against weaker
ones.

Another valuable byproduct of the two runs is that we got an estimate
of the bonus points for the second player that makes the perfect game a tie, 
that is, \emph{fair komi}. There are speculations on this
subject in the Go community and a general agreement that, with Chinese
scoring, a komi of 7.5 should be quite balanced.

\begin{figure}[ht]
  \caption{Evolution of the estimates $\hat\alpha_\varnothing$ and
    $\hat\beta_\varnothing$ of the initial empty board position for
    the two runs of \nine\ SAI. The value of $\hat\alpha_\varnothing$, above, should be
    interpreted as an estimate of the \emph{fair komi}, that is, the
    bonus points for the second player that make the perfect game a tie. The value of $\hat\beta_\varnothing$, below, increases indicating
    that the sigmoid corresponding to the initial position is becoming
    slowly steeper.}
	\label{f:alpha_beta}
	\begin{center}
		\includegraphics[width=.7\textwidth]{\dirfig/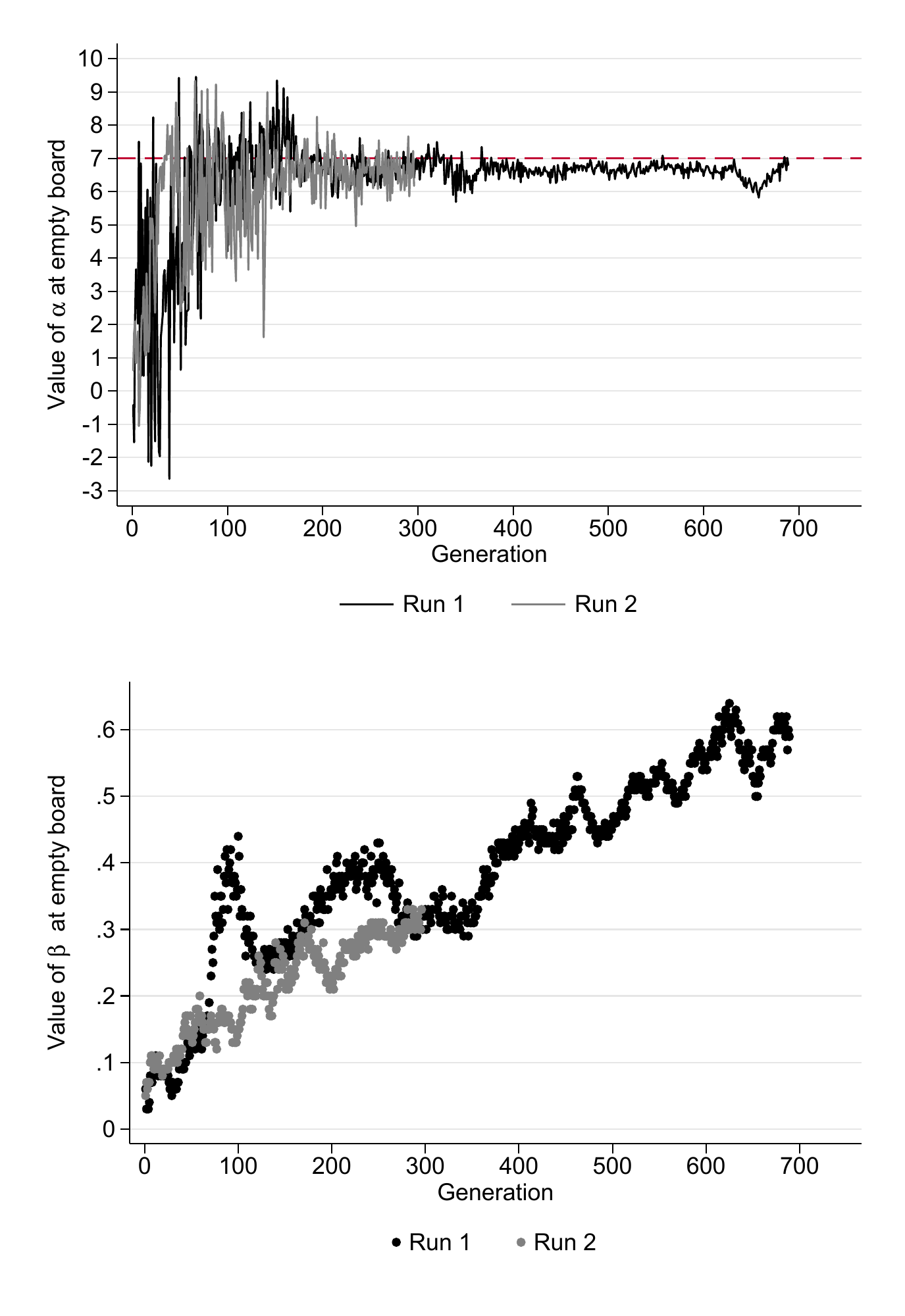}
	\end{center}
\end{figure}

Figure~\ref{f:alpha_beta} shows that SAI believes that the fair komi
should be 7 points.  This is confirmed by both runs, despite the fact
that the final networks of the two runs have different preferences for
the first moves: in the first run 4-4, 6-6, 5-6, 6-5; in the second
run 5-5, 3-2, then variable. These sequences evolved slowly along the
runs and the small oscillations in the fair komi estimate correspond
to the times when these moves preferences changed.

\subsection{Score in a Go game}
\label{s:scoring}

The final score of a Go game is a surprisingly delicate construct. The
Japanese and Chinese scoring methods are intrinsically difficult to
implement in an algorithm: when two human players agree that the game
is finished, they both pass and the score is then decided by manually
attributing stones and areas of the board. The Tromp-Taylor scoring
method on the other hand can be implemented quite efficiently, and is
in principle equivalent to the Chinese one, in the sense that an
unfinished game position is winning\footnote{Here we are referring to
  the concept of \emph{winning position} for abstract perfect
  information two-players zero-sum games, not to some condition on the
  expected winrate.} for Black under Chinese scoring if and only if it
is winning for Black under Tromp-Taylor scoring.

In fact, under Chinese rules, it is always possible to continue a
finished game, without changing the score, to a point when no more
moves can be played by either player without losing points, and at
that point the Tromp-Taylor score is equal to the Chinese score.
Basically, a game of Go is finished when both players agree on how
such a procedure would end.

For this reason Tromp-Taylor scoring was implemented in Leela Zero and
in SAI, and by that, mature networks learnt to estimate the winrate
and, in the case of SAI, also the expected score difference, which is
the parameter $\alpha$ with appropriate sign and modified by komi.

To avoid long final phases, the program resigns, when the winrate
estimate drops below a predefined threshold, and in fact self-play
games usually end by resignation of the losing side: pass is chosen
very rarely, because ending the game by double pass would trigger
Tromp-Taylor scoring that is typically not correct.  When a game is
decided by resignation, scoring is undefined, when it is decided by
double pass, there is no guarantee that Tromp-Taylor score is already
equal to the Chinese one. As a consequence, scoring is not a native
feature in Leela Zero or SAI.

For our experiments we needed a precise evaluation of the score of a
finished game, even if it ended by resignation at a fairly early
position. But it is difficult to implement an algorithm to compute
exactly the score of such a game, as it would in practice require to
conduct the game up to the point when Tromp-Taylor scoring can be
applied, and this leads to a painstakingly long final part, after the
winner is already decided.

\begin{figure}[ht]
  \caption{Estimate of the score at a position $s$ when Black resigned
    in a standard $7.5$ points komi game. The black sigmoid is the
    estimate of $\rho_s$ made by a strong net: based on this, the score from the point of view of Black would be $-1.97$, with a 90\% confidence
    interval of $[-5.96 ; -2.01]$. The gray sigmoids are the estimates
    of $\rho_r$ in positions of a subtree rooted on $s$ and generated
    by the same strong net with 1,000 visits. The final estimate of
    the score of the game is the median of such $\hat\alpha_r$'s. In this
    example, the estimate of the score from the point of view of Black is $-2.59$, which is very close
    to $-2.5$, the true score of the game, according to the expert evaluation of a strong player.  }
	\label{f:scoring}
	\begin{center}
		\includegraphics[width=.9\textwidth]{\dirfig/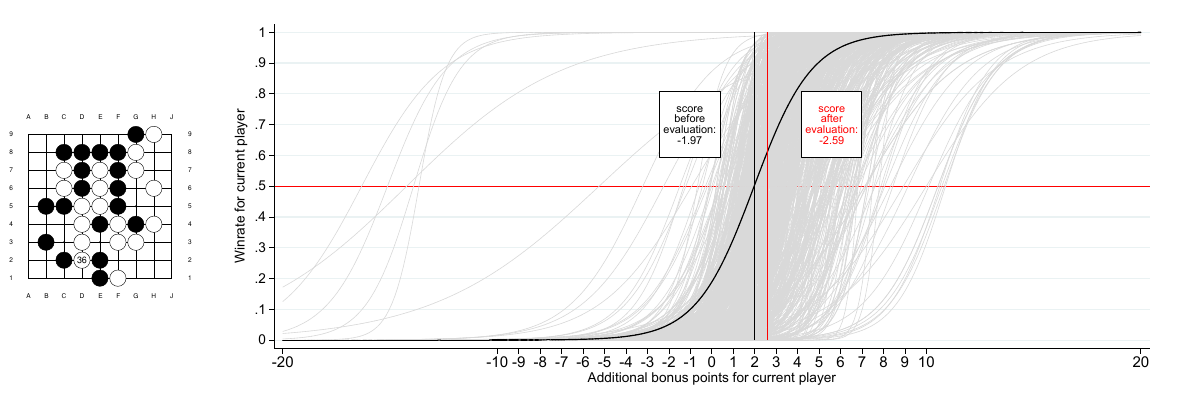}
	\end{center}
\end{figure}

In this work we exploited the features of SAI itself to estimate
score. The first choice could have been to have a standard strong SAI
network compute $\hat\alpha_s$ on the position $s$ when the loser
resigns. However we realized after some experiments that this estimate
is fairly unstable, in particular when $\hat\beta_s$ is low, and decided to
make it more precise, by aggregating information from a large sample
of positions in the subtree of visited nodes rooted at $s$. This
is the same principle introduced by AlphaGo Zero to assess the
winrate, but instead of the average, we chose the median, which proved
to be stable when based on 1,000 visits. The algorithm was validated
by an expert player on a set of 20 games. Figure~\ref{f:scoring} shows
an example.

\subsection{The experimental setting}\label{s:experimental_setting}

In order to conduct our experiments, we selected a set of strong and a
set of progressively weaker nets. We split the around 1,000 nets of
SAI (see Figure~\ref{f:elo}) in 20 quintiles of Elo. We randomly chose
3 nets, $S1$, $S2$ and $S3$, in the highest quintile stratified by
time in the training, in order to obtain strong nets qualitatively
different from each other. We then randomly chose one net in each of
the following strata: 14th, 8th, and 4th, and denoted them by $W1$
(the strongest), $W2$, and $W3$ (the weakest).  According to a
qualitative assessment, $W3$ is stronger than a strong amateur, but is
not at professional level.

To calibrate the nets, we had each of the strong nets play against
itself and against all the weak nets 100 times, half times with black
and half times with white, with komi 7.5 and with 1,000 visits.

\begin{figure}[ht]
	\caption{Calibration of the 3 strong nets versus themselves and each of the 3 weak nets.}
	\label{f:calibration}
	\begin{center}
		\includegraphics[width=.9\textwidth]{\dirfig/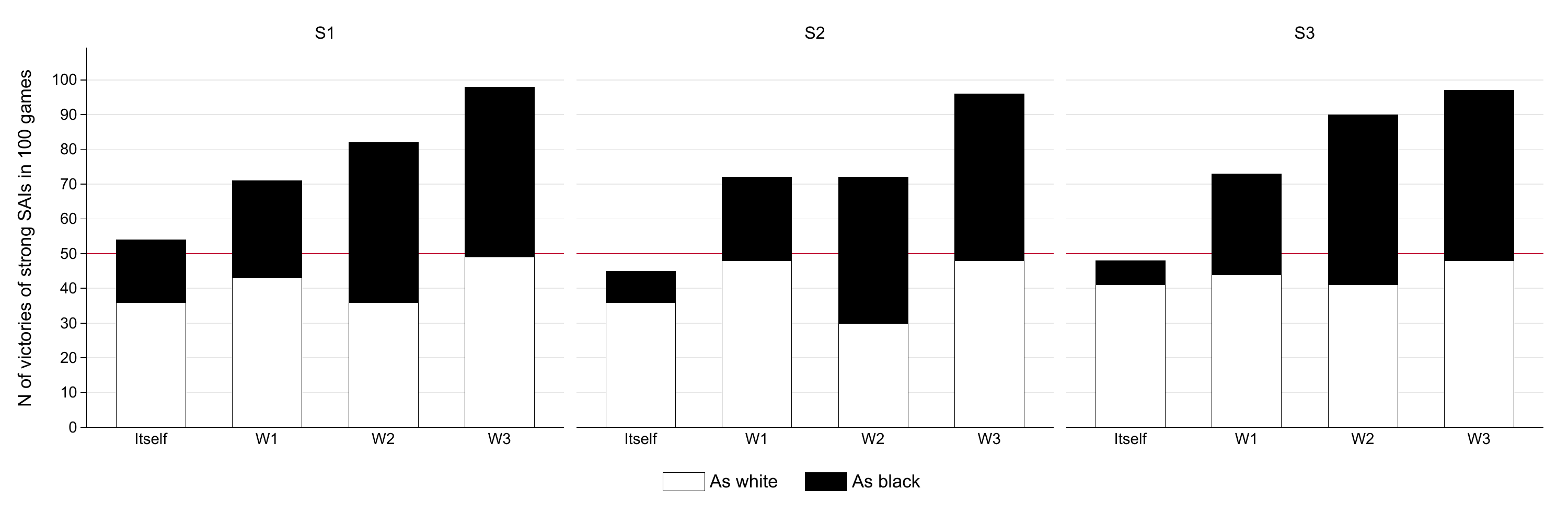}
	\end{center}
\end{figure}

The result is shown in Figure~\ref{f:calibration}. As expected, each
net won against itself around half of the times, although it would win
more often with white, consistently with the assessment that komi 7.5
is unbalanced in favor of White (see
Section~\ref{s:training_outcome}). The proportion of victories against
the weak nets progressed as expected, with $W3$ losing almost, but not
all, of the games against all three strong nets.

Since the results on 100 games showed a little too much variability, in the next experiments we played 400 games for each setting, in order to reduce the
uncertainty by half. The number of visits was kept to 1,000. The code of the experiment, as well as the files of the games in Smart Game Format, is available at the link~\cite{link_experiments_code_and_sgfs}, and the datasets containing the results underlying the figures is available at the link~\cite{datasets_csv}.

\clearpage

\section{Results}
\label{s:results}

\subsection{SAI can play with handicap}
\label{s:handicap}

We expected that, thanks to the knowledge of the sigmoid based on the
experience of branching, SAI was able to play with handicap, at least
to a reasonable extent. To prove this we conducted two experiments.

\begin{figure}[ht]
  \caption{Games of the 3 strong nets versus themselves and each of
    the 3 weak nets, with score handicap increasingly disadvantageous for the
    strong net (in relative terms with respect to komi 7.5). In the table, the percentages of victories are averages across the strong nets. The standard deviations (SD) are computed between the averages of the three nets.}
	\label{f:KOMI}
	\begin{center}
		\includegraphics[width=.9\textwidth]{\dirfig/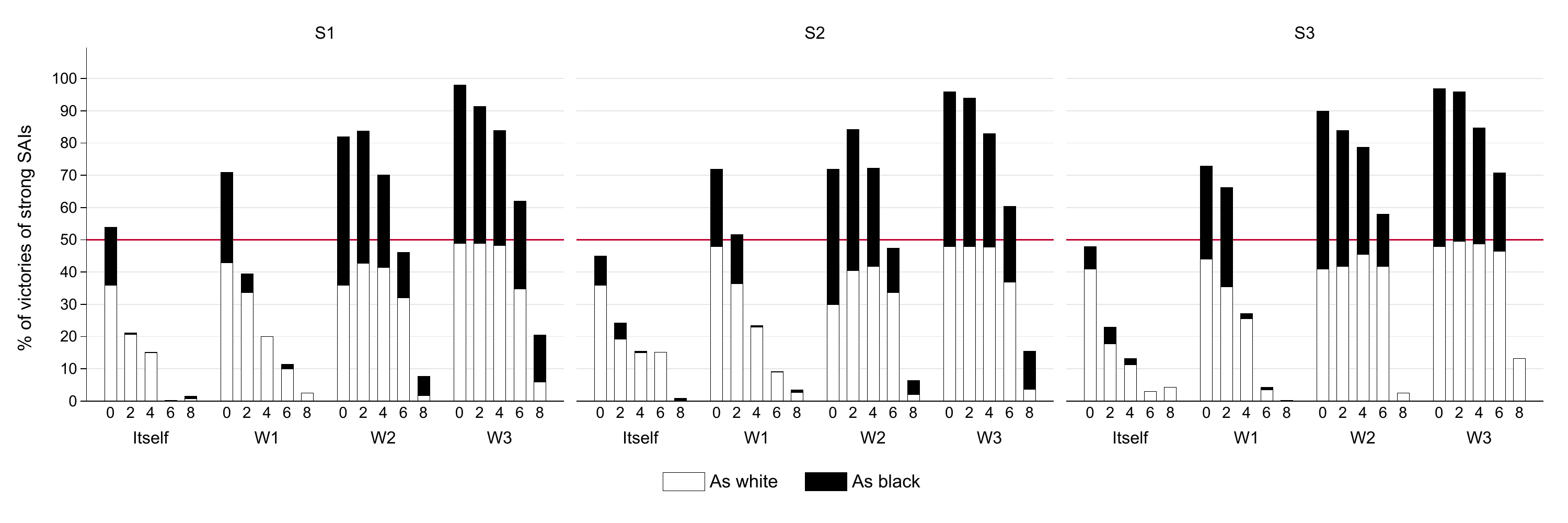}
		\begin{tabular}{llllllllll}
		&&\multicolumn{4}{c}{\bf Opponent}   \\ \cline{3-6}
		\bf Malus & \bf Color &\bf Itself &\bf W1 &\bf W2 &\bf W3 \\ \hline
		\input{\dirtab/KOMI_tab.tex} \hline
		\end{tabular}
	\end{center}
\end{figure}

In the first experiment, we had each of the strong nets play against
itself and against all the weak nets, half times with black and half
times with white, with an increasing score handicap, that is, with an increasing level of komi. The result is
shown in Figure~\ref{f:KOMI}, where the score handicap is represented in relative
terms with respect to an even game of 7.5, i.e. ``2'' means 5.5 if the
strong net is white, and 9.5 if the strong net is black.

When playing against themselves and against $W1$ as white, the strong
nets were able to win a sensible share of games with up to 6 points of score handicap, while as black their share of victories decreased to
less than 2\% on average at 4 points. Against the two weaker nets, all the
strong nets kept an average share of victories higher than 70\% with white, and a reasonably high share 
of victories with black (higher than 29\% on average), 
when playing with up to 6 points
of malus. With 8 malus points, the strong nets
could still win an appreciable share of times (5.6\% and 16.4\%), with both colors.

\begin{figure}[ht]
  \caption{Games of the 3 strong nets as white versus $W3$, with
    various levels of handicap. On first row, score handicap: H1 corresponds to komi 0.5. On second row, positional handicap: the starting board contains
    2 black stones and white plays first, and H2 corresponds to komi 0.5. Since
    the strong nets play white, lower komi points (from left to right)
    mean higher handicap. In the table, the percentages of victories are averages across the strong nets. The standard deviations (SD) are computed between the averages of the three nets.}
	\label{f:handicap}
	\begin{center}
		\includegraphics[width=.9\textwidth]{\dirfig/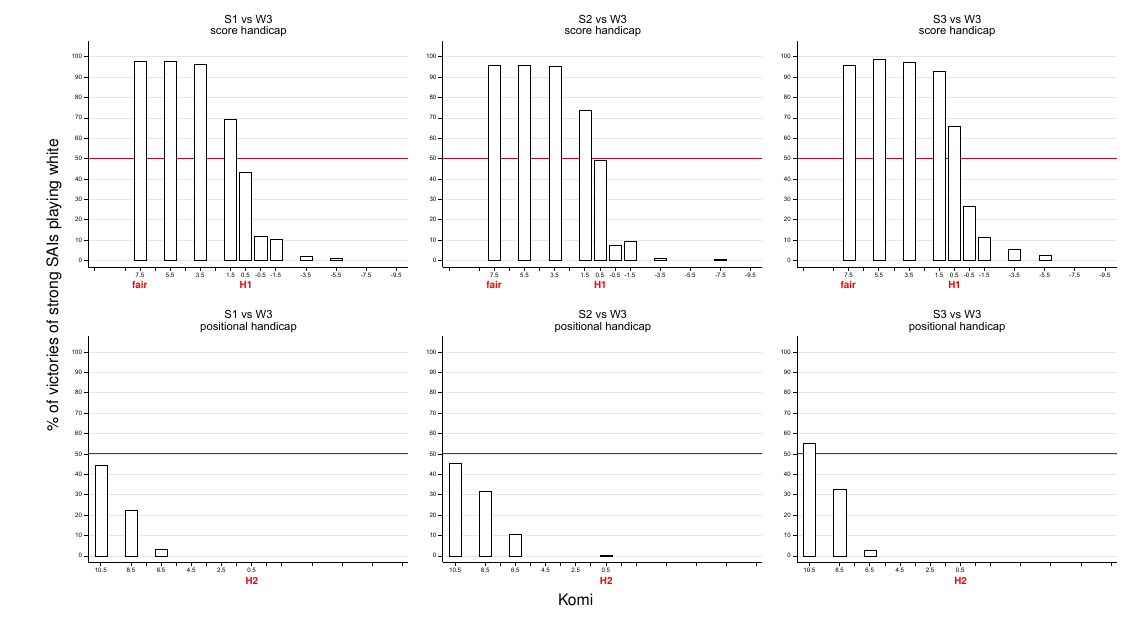}
		\tiny
		\begin{tabular}{lccl}
		\bf Type of handicap &\multicolumn{1}{p{1.5cm}}{\bf\centering Malus points}&\bf Komi &\multicolumn{1}{c}{\bf \%} \\\hline
		\input{\dirtab/stones_tab.tex} \hline
		\end{tabular}
	\end{center}
\end{figure}

In the second experiment, we pushed the handicap further, having the
three strong nets play against $W3$ both with additional, more extreme, score handicap and with the positional concept of handicap: traditionally, ``one handicap'' (H1) means having no
komi (actually, 0.5, to avoid ties), ``two handicaps'' (H2) means having white start the game
but with two black stones placed on the board. To follow the tradition, we had the three strong nets play only with white. H2 is considered to be prohibitive on a \nine\ board: to make it less
so, in the experiment we countered it with bonus points in favor of white. The
result is shown in Figure~\ref{f:handicap}.

The upper side of the figure integrates the results of the previous
experiment, showing that H1 (0.5 komi) is a handicap that makes games
between the strong nets and $W3$ even.

As expected, H2 proved prohibitive, even though in one single game
$S2$ was able to win. When H2 was played with 6.5 komi, in few cases
strong nets were able to recover, and with 8.5 the setting was
challenging again for $W3$. With 10.5 the game was again approximately
even.
\clearpage

\subsection{SAI can target high scores}\label{s:target_scores}

SAI nets can play with different agents, according to which of the
value functions introduced in Section~\ref{s:agent_model} is used. We expected that SAI, when
playing with agents with high parameters, due to the systematic
underestimation of the number of advantage points, would have targeted
high scores. We also expected that this would come at
the price of reducing its winning probability, due to two composite
effects: on the one hand, when playing against a strong opponent, SAI
would overlook more solid moves, towards moves riskier but with a
higher reward; on the other, when being in disadvantage, SAI would have the agent
provide a delusional overestimate of its chances of victory,
therefore jeopardizing its choices.

\begin{sidewaysfigure}[ht]
  \caption{Games of the 3 strong nets versus themselves and each of
    the 3 weak nets, with agents adopting different value functions
    parameterized by $\lambda$ and $\mu$. In each subfigure, the lower
    part shows the winning probability of the strong net, the upper
    part shows the average final score of the games won by the strong
    net.}
  \label{f:agent}
  \begin{center}
    \includegraphics[width=.99\textwidth]{\dirfig/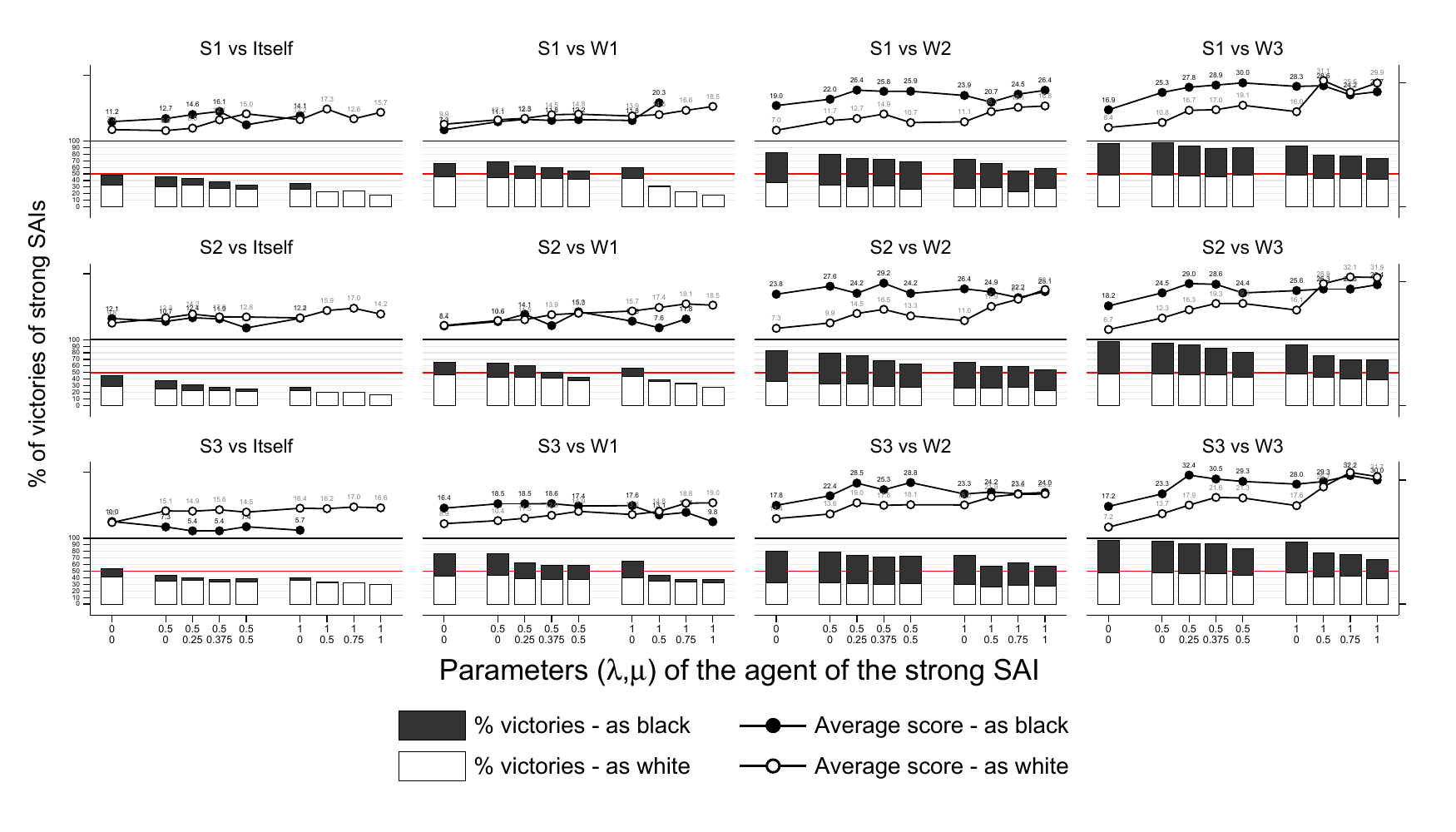}
  \end{center}
\end{sidewaysfigure}

To explore those effects, we had the 3 strong nets play against
themselves and against the 3 weak nets with a family of agents:
$\lambda$ equal to $0.5$ or $1$, and $\mu$ being in turn $0$,
$0.5\lambda$, $0.75\lambda$, and $\lambda$.
The result is shown in Figure~\ref{f:agent}.

\begin{sidewaystable}[ht]
  \caption{Data represented in Figure~\ref{f:agent}, averaged across strong nets. Average score is only computed when at least 5 games were won. The scores are averages of the baseline scores of the three strong nets when $\lambda=\mu=0$, and average differences with respect to baseline otherwise. The standard deviations (SD) are computed between the averages of the three nets.}
	\label{t:agent}
	\begin{center}
		\begin{tabular}{llllllllllllllllllllll}
		&&&\multicolumn{8}{c}{\bf Opponent}   \\ \cline{4-11}
		\bf Color &$\lambda$ & $\mu$ & \multicolumn{2}{c}{\bf Itself} & \multicolumn{2}{c}{\bf W1} & \multicolumn{2}{c}{\bf W2} & \multicolumn{2}{c}{\bf W3}  \\ 
		&&&\multicolumn{1}{c}{\bf  \%} &\multicolumn{1}{c}{\bf  Score}&\multicolumn{1}{c}{\bf  \%} &\multicolumn{1}{c}{\bf  Score}&\multicolumn{1}{c}{\bf  \%} &\multicolumn{1}{c}{\bf  Score}&\multicolumn{1}{c}{\bf  \%} &\multicolumn{1}{c}{\bf  Score} \\\hline
		Black&0&0&30.0 (SD: 5.0)&11.1 (SD: 1.1)&50.0 (SD: 17.3)&10.7 (SD: 5.0)&93.2 (SD: 2.0)&20.2 (SD: 3.2)&97.2 (SD: 1.0)&17.4 (SD: 0.6) \\ 
\cline{2-11} &0.5&0&22.5 (SD: 6.8)&-0.9 (SD: 2.1)&51.5 (SD: 12.6)&+2.7 (SD: 1.0)&93.5 (SD: 1.3)&+3.8 (SD: 0.8)&95.5 (SD: 0.9)&+6.9 (SD: 1.3) \\ 
&&0.25&15.3 (SD: 6.9)&-0.3 (SD: 4.0)&40.3 (SD: 6.3)&+4.3 (SD: 1.9)&86.2 (SD: 0.8)&+6.2 (SD: 5.3)&90.7 (SD: 0.6)&+12.3 (SD: 2.5) \\ 
&&0.375&12.5 (SD: 7.0)&+0.1 (SD: 4.7)&30.5 (SD: 12.8)&+2.3 (SD: 2.2)&81.8 (SD: 2.8)&+6.6 (SD: 1.1)&85.3 (SD: 4.3)&+11.9 (SD: 1.4) \\ 
&&0.5&10.3 (SD: 3.5)&-2.9 (SD: 1.5)&26.2 (SD: 16.1)&+4.3 (SD: 3.0)&80.3 (SD: 8.1)&+6.1 (SD: 5.3)&79.0 (SD: 5.0)&+10.5 (SD: 3.7) \\ 
\cline{2-11} &1&0&11.5 (SD: 4.8)&-0.4 (SD: 3.6)&36.2 (SD: 13.0)&+2.7 (SD: 1.7)&85.2 (SD: 5.8)&+4.3 (SD: 1.5)&89.7 (SD: 3.1)&+9.8 (SD: 2.1) \\ 
&&0.5&0.5 (SD: 0.5)&&8.5 (SD: 8.3)&+3.0 (SD: 8.8)&68.0 (SD: 5.8)&+3.1 (SD: 2.9)&70.2 (SD: 4.3)&+10.6 (SD: 2.1) \\ 
&&0.75&0.3 (SD: 0.6)&&3.7 (SD: 3.5)&+0.7 (SD: 3.8)&65.5 (SD: 3.0)&+3.2 (SD: 4.2)&63.5 (SD: 4.4)&+10.1 (SD: 4.2) \\ 
&&1&0.5 (SD: 0.9)&&3.7 (SD: 5.1)&-6.6&62.3 (SD: 0.3)&+5.0 (SD: 3.2)&60.2 (SD: 2.0)&+10.6 (SD: 2.0) \\ 
\hline White&0&0&69.2 (SD: 13.5)&8.9 (SD: 1.3)&90.2 (SD: 4.9)&9.2 (SD: 0.7)&70.5 (SD: 4.4)&8.6 (SD: 2.4)&97.2 (SD: 0.8)&7.5 (SD: 0.8) \\ 
\cline{2-11} &0.5&0&61.8 (SD: 9.8)&+2.5 (SD: 3.1)&88.0 (SD: 0.5)&+2.0 (SD: 0.5)&65.8 (SD: 0.8)&+3.2 (SD: 1.3)&96.8 (SD: 1.0)&+4.8 (SD: 2.1) \\ 
&&0.25&61.7 (SD: 14.4)&+3.5 (SD: 2.5)&84.3 (SD: 4.6)&+2.8 (SD: 0.1)&63.3 (SD: 1.9)&+6.8 (SD: 1.0)&94.5 (SD: 0.0)&+9.5 (SD: 1.2) \\ 
&&0.375&56.5 (SD: 11.0)&+4.6 (SD: 1.6)&82.8 (SD: 6.0)&+4.6 (SD: 0.6)&60.3 (SD: 3.3)&+7.8 (SD: 1.4)&93.8 (SD: 1.3)&+11.8 (SD: 3.0) \\ 
&&0.5&55.0 (SD: 11.6)&+5.2 (SD: 2.3)&78.0 (SD: 5.2)&+5.6 (SD: 0.6)&56.8 (SD: 4.5)&+5.5 (SD: 1.6)&90.8 (SD: 5.0)&+12.5 (SD: 1.7) \\ 
\cline{2-11} &1&0&57.3 (SD: 14.3)&+4.7 (SD: 2.2)&85.0 (SD: 4.4)&+5.1 (SD: 1.6)&57.2 (SD: 2.5)&+4.8 (SD: 1.6)&96.5 (SD: 0.5)&+9.1 (SD: 1.4) \\ 
&&0.5&50.8 (SD: 13.3)&+7.6 (SD: 2.1)&68.7 (SD: 7.3)&+6.4 (SD: 2.0)&55.3 (SD: 3.7)&+10.0 (SD: 0.9)&85.2 (SD: 1.9)&+21.4 (SD: 1.7) \\ 
&&0.75&52.0 (SD: 12.6)&+6.6 (SD: 1.2)&59.7 (SD: 11.5)&+9.0 (SD: 2.0)&52.7 (SD: 6.3)&+12.3 (SD: 1.6)&85.0 (SD: 2.6)&+23.0 (SD: 5.1) \\ 
&&1&43.8 (SD: 14.1)&+6.6 (SD: 2.0)&52.5 (SD: 14.7)&+9.5 (SD: 0.8)&52.5 (SD: 4.8)&+14.1 (SD: 4.0)&80.8 (SD: 3.2)&+23.7 (SD: 1.9) \\ 
 \hline
		\end{tabular}
	\end{center}
\end{sidewaystable}

As expected, the winning probability of the strong nets when playing
against themselves and against $W1$ decreased rapidly with increasing
$\lambda$, especially for high values of $\mu$ and when playing as
black. As expected, however, for white the average score when winning, 
which was around $10$ points at baseline with $\lambda=\mu=0$,  
showed a
remarkable improvement, up to $+7.6$ points on average when playing 
 against itself with $\lambda=1$ and $\mu=0.5$, and up to $+9.5$ points 
on average when playing  against $W1$ with $\lambda=\mu=1$.

The loss of strength against $W2$ and $W3$ was not so apparent, even
for high values of $\lambda$ and $\mu$. In particular when playing
white against $W3$, the strong nets experienced a very small
loss of stregth. On the other hand, the improvement in score was substantial. 
Against $W2$, when playing black, from a baseline 20.2 the average improvement was up to $+6.6$ with $(\lambda,\mu)=(0.5,0.375)$, 
and when playing white, from a baseline $8.6$ the average improvement was up to $+14.1$ points 
with $\lambda=\mu=1$. Against $W3$, when playing black, from a baseline $17.4$ the average improvement was up 
to $+12.3$ with $(\lambda,\mu)=(0.5,0.25)$, and when playing white, from a baseline $7.5$ the average improvement was up to $+23.7$ points 
with $\lambda=\mu=1$.

The potential of the
family of value functions 
would be better exploited if the choice of the value function was
targeted to the situation in the game, for instance the winning probability
and the strength of the opponent. 
We applied a variable agent by having the strong nets play
against themselves and against $W3$ 
using three agents ($(0.5,0)$, $(1,0)$
and $(1,0.5)$) which were only activated if the winrate was higher
than a threshold of, in turn, 50\% and 70\%.

\begin{figure}[ht]
	\caption{Games of the 3 strong nets versus themselves, with agents adopting different value functions parameterized by $\lambda$ and $\mu$, when the pointwise estimate of victory is above a pre-defined threshold of 0\%, 50\% and 70\%. In each subfigure, the lower
    part shows the winning probability of the strong net with variable agent, the upper
    part shows the average final score of the games won by the strong
    net. In the table, the percentages of victories are averages across the strong nets, and the scores are averages of the baseline scores of the three strong nets when $\lambda=\mu=0$, and average differences with respect to baseline otherwise. Average score is only computed when at least 5 games were won. The standard deviations (SD) are computed between the averages of the three nets.}
	\label{f:variable_agent}
	\begin{center}
		\includegraphics[width=.9\textwidth]{\dirfig/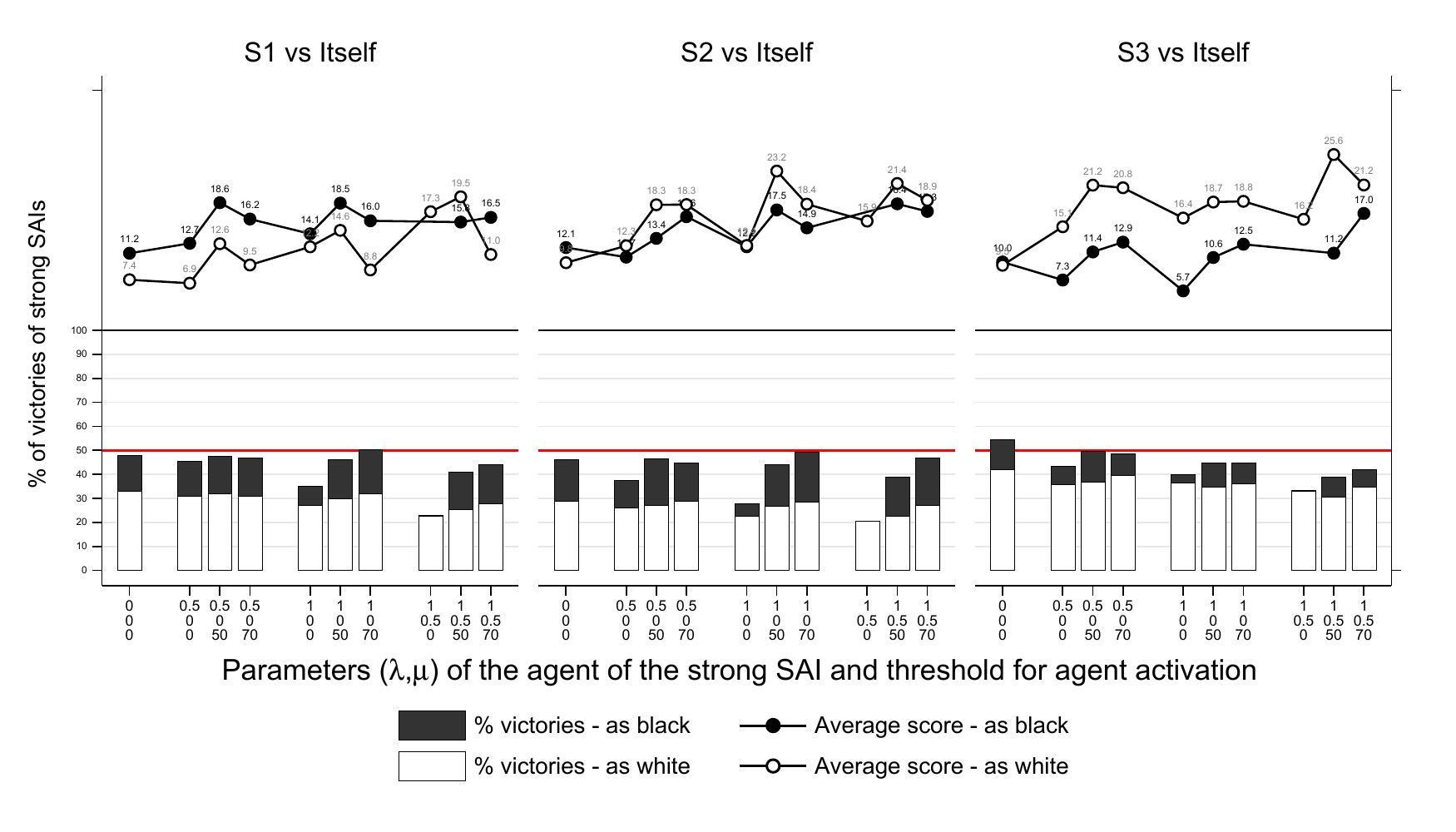}
		\begin{tabular}{lllllllllll}
		&&&&\multicolumn{2}{c}{\bf Opponent}   \\ \cline{5-6}
		\bf Color &$\lambda$ & $\mu$ & \multicolumn{1}{c}{\bf Threshold} & \multicolumn{2}{c}{\bf Itself} \\ 
		&&&&\multicolumn{1}{c}{\bf \%} &\multicolumn{1}{c}{\bf Score} \\ \hline 
		\input{\dirtab/variable_agent_tab.tex} \hline
		\end{tabular}
	\end{center}
\end{figure}

The result of the games between the strong nets and themselves is shown in Figure~\ref{f:variable_agent}. The threshold
effectively eliminated or contained the loss in strength. With black the winning probability, which was $30.0\%$ on average at baseline,
was not affected or even increased for $\mu=0$ and had a lowest point of $26.8\%$ with $\mu=0.5$. With white the baseline winning probability was $69.2\%$ on average, especially driven by $S3$, and only decresed below $60.0\%$ on average for $(\lambda,\mu)=(1,0.5)$ and threshold $50\%$.
The gains in
score were maintained or increased with thresholds. With black, from a baseline of $11.1$ on average, the gain was $+5.9$ on average with $(\lambda,\mu)=(1,0.5)$ and threshold $70\%$. With white, from a baseline of $8.9$ on average, the gains were higher with threshold $50\%$, up to $+13.3$ on average with $(\lambda,\mu)=(1,0.5)$ (mostly driven by $S3$, and with a decreased winning probability). With threshold 70\% the gains were smaller but still sustained, up to $+8.2$ on average with $(\lambda,\mu)=(1,0.5)$.

\begin{figure}[ht]
	\caption{Games of the 3 strong nets versus $W3$, with agents adopting different value functions parameterized by $\lambda$ and $\mu$, when the pointwise estimate of victory is above a pre-defined threshold of 0\%, 50\% and 70\%. In each subfigure, the lower
    part shows the winning probability of the strong net with variable agent, the upper
    part shows the average final score of the games won by the strong
    net. In the table, the percentages of victories are averages across the strong nets, and the scores are averages of the baseline scores of the three strong nets when $\lambda=\mu=0$, and average differences with respect to baseline otherwise. The standard deviations (SD) are computed between the averages of the three nets.}
	\label{f:variable_agent_weakopponent}
	\begin{center}
		\includegraphics[width=.9\textwidth]{\dirfig/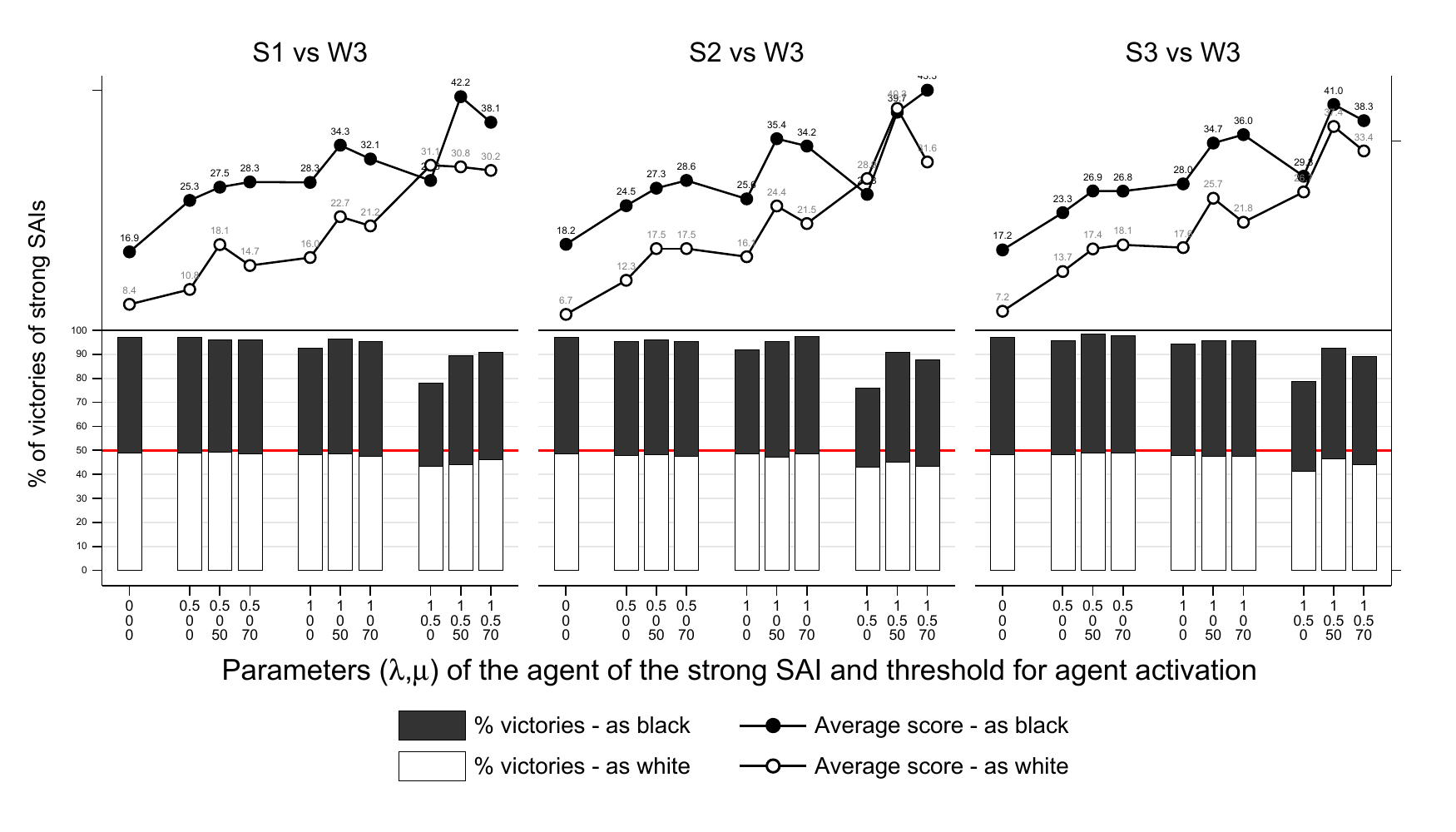}
		\begin{tabular}{lllllllllll}
		&&&&\multicolumn{2}{c}{\bf Opponent}   \\ \cline{5-6}
		\bf Color &$\lambda$ & $\mu$ & \multicolumn{1}{c}{\bf Threshold} & \multicolumn{2}{c}{\bf W3} \\ 
		&&&&\multicolumn{1}{c}{\bf \%} &\multicolumn{1}{c}{\bf Score} \\ \hline 
		\input{\dirtab/variable_agent_weakopponent_tab.tex} \hline
		\end{tabular}
	\end{center}
\end{figure}

The result of the games between strong nets and $W3$  is shown in Figure~\ref{f:variable_agent_weakopponent}. 
The baseline winning rate had not been bothered by $\lambda=0.5$, but was hampered with $\lambda=1$ and both $\mu$ for black, and for  $(\lambda,\mu)=(1, 0.5)$ for white: both thresholds restored it to a level equal or slightly lower than the baseline. From the baseline of 17.4 for black and 7.5 for white, the score peaked with threshold $50$, up to +23.5 for black and +28.7 for white, in both cases with $(\lambda,\mu)=(1, 0.5)$. Threshold $70$ obtained consistently less remarkable performances.

\clearpage

\subsection{SAI can recover from very severe disadvantage}\label{s:dramatic}

\begin{figure}[ht]
  \caption{Games of the 3 strong nets versus the weakest net, playing
    white with handicap incremented or decreased. Since
    the strong nets play white, lower komi points mean higher
    disadvantage.  Type of handicap 1: score handicap. Type of handicap 2: positional handicap (the starting board contains 2 black
    stones and white plays first). In the table, the percentages of victories are averages across the strong nets. The standard deviations (SD) are computed between the averages of the three nets.}
	\label{f:dramatic}
	\begin{center}
		\includegraphics[width=.9\textwidth]{\dirfig/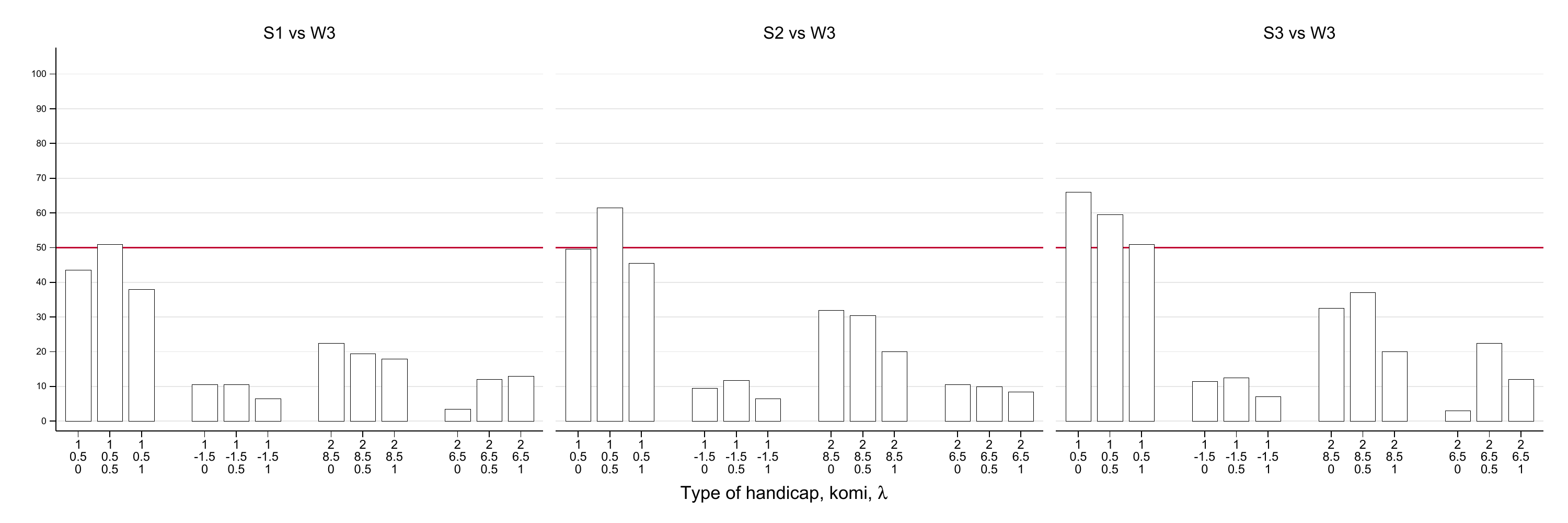}
		\begin{tabular}{llll}
		\multicolumn{1}{p{2cm}}{\bf \centering Type of handicap} &\multicolumn{1}{c}{\bf  Komi} & \multicolumn{1}{c}{\bf  $\lambda$} &\multicolumn{1}{c}{\bf \%} \\ \hline
		\input{\dirtab/agent_onStrongWithHandicap_tab.tex} \hline
		\end{tabular}
	\end{center}
\end{figure}

As demonstrated in Section~\ref{s:handicap}, SAI can play with
handicap, and strong nets maintain a remarkably high winning probability even when they play with substantial disadvantage. Victory
becomes rare in H1 and H1 with additional malus points; or
H2 with 6 or even 8 bonus points. In such severe situations we
investigated whether having the strong net overestimate its advantage
would help it keeping a solid game until the weak net made some error,
allowing the strong net to leverage on its own superiority and win. 
To this aim we had the 3 strong nets
play with $W3$ in the four severe disadvantageous situations, with
$\lambda$ set to $0.5$ and to $1$.
The result is shown in Figure~\ref{f:dramatic}.

With $\lambda=0$, both $S1$ and $S3$ had won less than $5\%$ of the
times with 2 handicaps and a bonus of 6 points (komi 6.5). In both
cases, setting $\lambda$ to $0.5$ increased substantially the winning probability, to $12.0\%$ and $22.5\%$ respectively. In the case of
$S2$, that had had a better performance than $S1$ and $S3$ in this
extreme situation, increasing $\lambda$ to $0.5$ didn't have any
noticeable effect. Setting $\lambda$ to 1 did not further improve the
winning probability for any of the strong nets. In the other, less
extreme situations of disadvantage, the effect of setting $\lambda$ to
$0.5$ was inconsistent, while further increasing $\lambda$ to $1$
never improved the winning probability.

\clearpage

\subsection{SAI can minimize suboptimal moves}
\label{s:suboptimal}

As mentioned in the Introduction, AlphaGo playing suboptimal moves and winning often by a small margin is common knowledge among Go players, based on professional analysis of public AlphaGo games against humans\cite{invisible}. In order to quantify this claim and, at the same time, prove that SAI acts less suboptimally, in July 2019 the following experiment was organised. Since AlphaGo and AlphaGo Zero are not publicly available, we selected a recent, very strong Leela Zero (LZ) net.\footnote{See \url{https://github.com/leela-zero/leela-zero/issues/863\#issuecomment-497599672}.} We ran 200 games between LZ and W3, the weakest net in our experiments. LZ was assigned white and won all the games. We drew a 10\% random sample from the 200 games. Our strongest net, S1, had won 194 times out of 200 games played as white against W3, with $(\lambda,\mu)=(1,0)$ (see Section~\ref{s:target_scores}). We drew a random sample of 20 games from the 194. We shuffled the 40 games and labeled them with an anonymous identifier. 

We asked two strong amateur players (4D and 3D) to score the 40 games and to identify whether the winner had played suboptimal moves. We also asked them to rate their own estimate of score as `reliable' or `non reliable'. The two assessors did not communicate with each other during assessment. As a result, the scores of 32 games were rated `reliable' by both assessors. The average difference in score between the two assessors was 1.53, in detail 0.8 among `reliable' and 4.3 among `non reliable' games. We computed the mean of the two scores and we linked the data to the identity of the winners: LZ or SAI. We found that the average scores of LZ and SAI were, respectively, 6.3 and 16.0 (single-tail $t$-test: $p<0.001$), or 6.0 and 15.0 when restricting to `reliable' games (single-tail $t$-test: $p=0.006$). The games with no suboptimal moves were 18 (90\%) for SAI and 11 (55\%) for LZ ($\chi^2$ test: $p=0.013$). The files of the games in Smart Game Format, the manual assessment and analysis are available at this link~\cite{link_validation}. In summary, even though LZ won more games, it got signficantly lower scores, and made significantly more suboptimal moves, with respect to SAI playing with an agent targeting high scores.

\subsection{SAI is superhuman}
\label{s:superhuman}

%
\subsubsection{SAI has professional level}

A match was scheduled on May 2019 between SAI and Hayashi Kozo 6P, 
a professional
Japanese player. The match was composed by 3 games, at alternate
colors, and komi $7.5$. The first to play black was to be chosen
randomly. SAI was set to play with net $S1$, $50,000$ visits,
$\lambda=\mu=0$, two threads, and resign threshold set at $5\%$.

\begin{figure}[ht]
	\caption{The sequence that would have led White to win in Game 1 between SAI and Hayashi Kozo 6P.}
	\label{f:amazing_sequence1}
	\begin{center}
		\includegraphics[width=.4\textwidth]{\dirfig/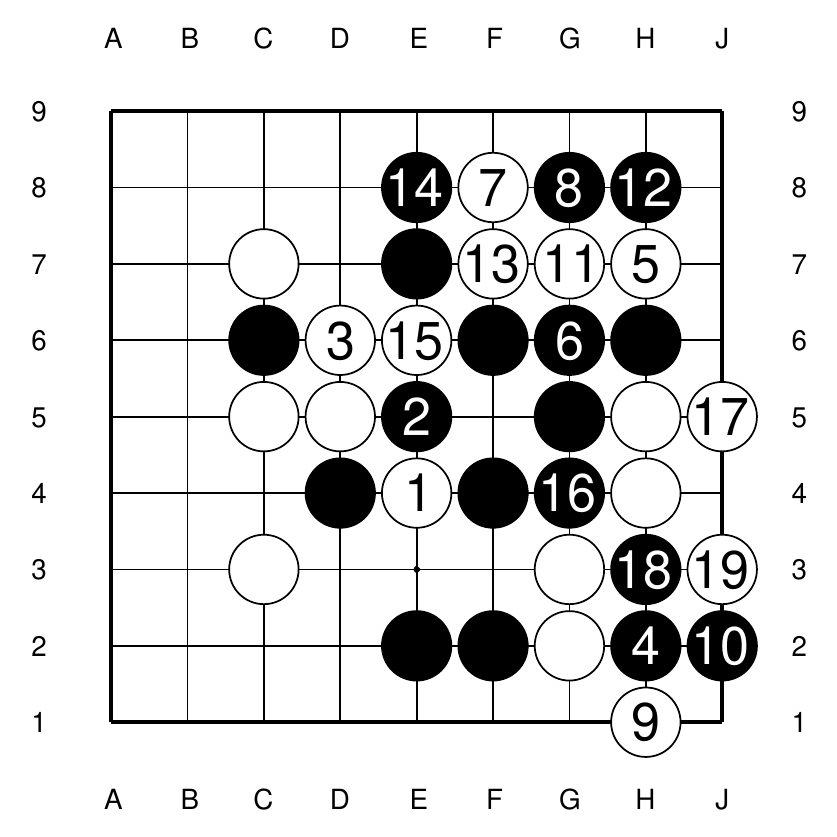}
	\end{center}
\end{figure}

\begin{figure}[ht]
	\caption{The sequence that would have led White to win in Game 3 between SAI and Hayashi Kozo 6P.}
	\label{f:amazing_sequence2}
	\begin{center}
		\includegraphics[width=.4\textwidth]{\dirfig/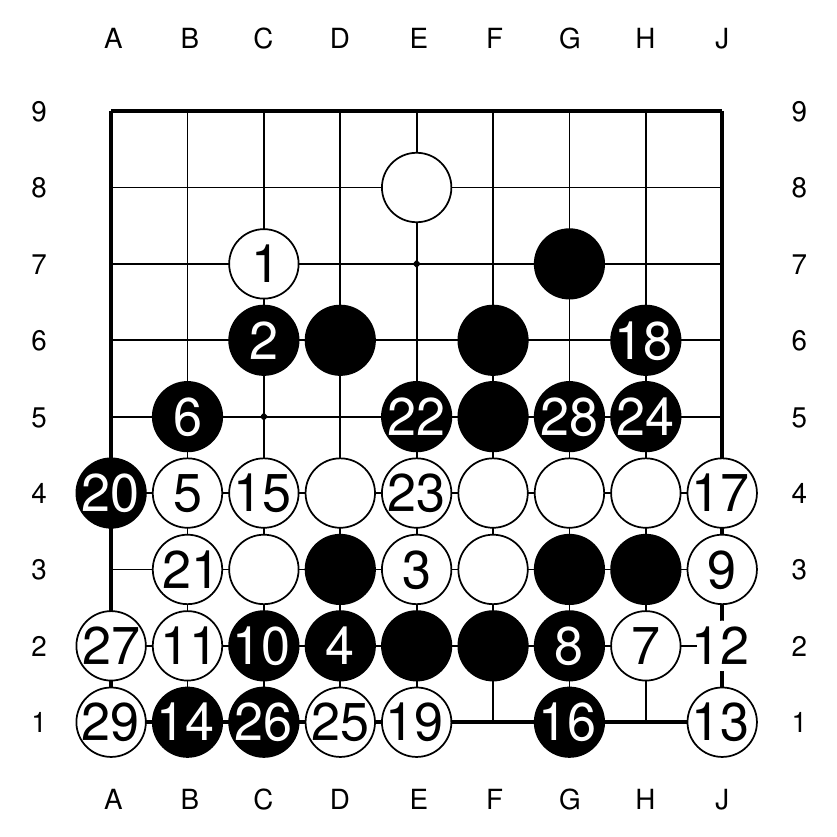}
	\end{center}
\end{figure}

\begin{description}

\item[Game 1. SAI (Black) vs Hayashi Kozo 6P (White)]
Due to the komi, White started with a lead, that SAI quantifies as approximately 62\% of winning probability. White managed to keep his advantage till move 18 where he played at H2, a regular move for a strong human player. However, following this move, SAI's evaluation of its own chances rose immediately to 86\%. The next sequence led to an easy win for Black, and we suspect that after move 27 SAI's choices started to be suboptimal (these games were played with $\lambda=\mu=0$, see Section~\ref{s:dev}). The game is represented in Appendix~\ref{s:games_hayashi}, Figure~\ref{f:games_hayashi_1}. A closer look into SAI's calculation revealed a sequence starting at move 18 (E4 E5 D6 H2 H7 G6 F8 G8 H1 J2 G7 H8 F7 E8 E6 G4 J5 H3 J3 J4 D8 F9 J3) that would have led White to win, consistently with his initial advantage. This sequence, represented in Figure~\ref{f:amazing_sequence1}, is truly extraordinary and an expert Go player judged it as superhuman.
\item[Game 2: Hayashi Kozo 6P (Black) vs SAI (White)]
The game was decided at an early stage by move 7, F6, which was judged by SAI as a mistake with respect to move G5. SAI managed this initial lead without giving Black any chance to recover. Eventually White won by 6.5 points. The game is represented in Appendix~\ref{s:games_hayashi}, Figure~\ref{f:games_hayashi_2}.
\item[Game 3: SAI (Black) vs Hayashi Kozo 6P (White)]
Game 3 was perfectly balanced until move 16, when White played B4. This move led to the forcing sequence C5 C4 C2 B2 E3 E4 that guaranteed life to the black group at the bottom,  as well as the initiative. After that Black played C7, leaving to White no choice but to resign. The game is represented in Appendix~\ref{s:games_hayashi}, Figure~\ref{f:games_hayashi_3}.  Instead of B4, SAI would have recommended another superhuman sequence, C7 C6 E3 D2 B4 B5 H2 G2 J3 C2 B2 J2 J1 B1 C4 G1 J4 H6 E1 A4 B3 E5 E4 H5 D1 C1 A2 G5 A1 F5, that would have led white to win the capturing race against the black group at the bottom. This sequence is represented in Figure~\ref{f:amazing_sequence2}.

\end{description}

\subsubsection{SAI won against a professional-level Go player with 6 additional malus points}

A match was scheduled on May 2019 between SAI and Oh Chimin 7D, a 7-Dan Korean
amateur player whose strength is estimated by the official Go rating
system of the European Go Federation GoR \cite{EGD} at 2,752 points, equivalent to
professional players: for instance Fan Hui 2P, the Chinese professional who contributed 
to the development of AlphaGo, has GoR 2,771.

The first game was set to be an even game, that is, komi 7.5; SAI was
scheduled to play white. The sequence was scheduled to run for 5
games; whenever SAI would won, the next game would be again with same
colors, and additional 2 malus points for SAI; whenever Oh Chimin 7D would
won, the next game would be with the same level of malus points for
SAI, and exchanged colors. SAI was set to play with net $S1$, $50,000$
visits, $\lambda=\mu=0$, two threads, and resign threshold set at
$5\%$.

The result of the match is summarized in Table~\ref{t:HCM}. SAI won
the first 3 games playing white. The fourth game was won by Oh Chimin 7D,
playing black with 6 additional bonus points, that is, komi was
$7.5-6=1.5$. As scheduled, in the next game SAI played with black and
the komi was set with same level of disadvantage: since SAI was black,
it was set to $7.5+6=13.5$. This game was won by SAI.

\begin{table}[ht]
	\caption{Result of the games of the match between SAI and Oh Chimin 7D, with increasing handicap for SAI.}
	\label{t:HCM}
	\begin{center}
		\begin{tabular}{ccrclc}
			\hline
			\bf Game&\multicolumn{1}{p{1cm}}{\bf \centering Color of SAI}&\bf Komi&\multicolumn{1}{p{1cm}}{\bf \centering Malus of SAI}&\bf Result&\bf Winner\\ \hline
			1&W&7.5&0&W+2.5&SAI\\
			2&W&5.5&2&W+R&SAI\\
			3&W&3.5&4&W+R&SAI\\
			4&W&1.5&6&B+R&OC\\
			5&B&13.5&6&B+1.5&SAI\\ \hline
		\end{tabular}
	\end{center}
\end{table}

\begin{description}

\item[Game 1. Oh Chimin 7D (Black) vs SAI (White)]

No significant mistake by Black could be noticed, according to SAI evaluation. The game was won by White by 2.5 points. The game is represented in Appendix~\ref{s:games_OC}, Figure~\ref{f:games_OC_1}. Since we estimated the fair komi to be 7, we can say that during the game Black progressively lost approximately 2 points. However, a thorough analysis of SAI evaluations suggests that White started playing suboptimal (these games were played with $\lambda=\mu=0$, see Section~\ref{s:dev}) moves starting from move number 20. This behaviour, similar to the one observed in Game 1 against Hayashi Kozo 6P, was indirectly confirmed by the next games, where SAI was able to win with up to 6 points of handicap.

\item[Game 2: Oh Chimin 7D (Black) vs SAI (White)]

White started with 2 malus points, i.e.~komi 5.5. According to this initial disadvantage, SAI's evaluation started below 50\% and remained that way up to move 11, when  Black played G5, instead of the move B7 that SAI would have recommended. At this stage the game was already decided in White's favor. White maintained the lead till Black's resignation. SAI's evaluation suggests that White was ahead of 4.5 points. The game is represented in Appendix~\ref{s:games_OC}, Figure~\ref{f:games_OC_2}. 

\item[Game 3 and 4: Oh Chimin 7D (Black) vs SAI (White)]
 
Game 3 and 4 were played with komi respectively 3.5 and 1.5 (i.e.~4 and 6 bonus points for Black). When  Black resigned at the end of  Game 3, he evaluated that he would have won with 2 more bonus points. Therefore, in the next game, Black replicated up to move 15, when White diverged from the Game 3. SAI evaluated its position slightly more disadvantageous in Game 4 compared to Game 3, where it suffered from additional 2 malus points. At move 16, in Game 3 it preferred D2, while B4 was its choice in Game 4. This is a corroboration that SAI's evaluations are komi-dependent. At move 16, SAI evaluated that both games were in an unfavorable position. In Game 3, Black missed the winning sequence at move 30 (the sequence C8 D1 H7 E1 F2 C1 C4 B5 B4 A4 J6 J3 B7 D5 A3 A5 C6 H1 D6 A2) that would have led him to win the capturing race against White's group, and eventually the game by 1.5 points. In Game 4, Black did not missed the chance and maintained the lead throughout the game (more than $50\%$ of winning probability), eventually winning by resignation. The games are represented in Appendix~\ref{s:games_OC}, Figure~\ref{f:games_OC_3} and Figure~\ref{f:games_OC_4}. 

\item[Game 5: SAI (Black) vs Oh Chimin 7D (White)]
 This time SAI took Black and 6 malus points, corresponding to a 13.5 komi. In the previous game the difference between SAI and Oh Chimin 7D was around 4/6 points, therefore we expected a well balanced game. Indeed, White managed to keep his initial lead until move 26 at H4. At this point, SAI evaluation of its own chances  rose from 33.8\% to 64.7\%. This decided the game in Black's favour: eventually Black won by 1.5 points. The game is represented in Appendix~\ref{s:games_OC}, Figure~\ref{f:games_OC_5}.  SAI, instead, spotted the winning sequence (F7 E8 H6 H5 F8 H8 J6 F9 J3 J5 D5 E6 G9 F6 E7 G7 E9 D8 D6 G8 B1 F4 H4) that would have led White to a 0.5 victory.

\end{description}
  
According to expert Go knowledge, winning against a professional-level player with 6 points of handicap on a \nine\ board is an achievement that classifies SAI as superhuman.


\section{Developments}
\label{s:dev}

\subsection{SAI and the game of Go}

Many opportunities associated with SAI's framework are still
unexplored. For instance, the variable agents introduced at the end of Section~\ref{s:target_scores} may be used in dependance of other parameters with respect to the pointwise estimate of the winrate, or used in conjunction with an estimate of the strength of the opponent.
Symmetrically, the ability of SAI to recover from very severe disadvantage against a weaker opponent, that was highlighted in Section~\ref{s:dramatic}, may probably be better exploited if the agent were variable. Indeed, it may be suggested that, as soon as a mistake from the weaker opponent re-opens the game, a variable agent could restore its proper assessment of the winrate, thus switching its aim from recovering points to targeting victory and improving effectively its chances. 

Moreover, note that SAI is able to win with substantial handicap, even in situations when the information stored in the sigmoid is of little help because the winrate is very low. Similarly, a SAI net is able to win when playing with $\lambda=\mu=1$, when the winrate is constantly close to $0.5$, even when playing with an equally strong opponent (itself). We speculate that both those this abilities amount to the quality of the \emph{policy}, that stems from experience gained during branched games. It would be interesting to test this hypothesis.


In the games we organized with human players, see Section~\ref{s:superhuman}, our primary goal was victory. For this reason we used $\lambda=\mu=0$. As a consequence, we observed that SAI made some suboptimal moves. A promising development is using SAI with variable $\lambda$ and $\mu$ against humans, for competitive and teaching purposes, in order for it to choose optimal moves with humans even when it is winning.

All such opportunities will have a chance to be fully exploited when SAI is trained to play on the \nineteen\ board.

\subsection{SAI and Deep Reinforcement Learning}


After the seminal papers on Atari~\cite{mnih2015humanlevel} and AlphaGo~\cite{AlphaGo}, Deep Reinforcement Learning (DRL) has been a major research topic. The SAI framework, at its essence, is a variation into the high-level, domain-independent aspects of DRL, stemming from the assumption that probability of success is a function of the targeted score, belonging to a parametric family of functions whose parameters can be learned by a neural network. The only requirement for SAI to contribute to an application is that success is linked to some score in the first place. This is true in many instances of zero-sum two-player games, for instance Othello, where a Leela Zero approach has been advocated for but not yet pursued~\cite{DBLP:journals/tciaig/LiskowskiJK18}. In fact, DRL was recently used to address broader challenges, such as multi-player games, with incomplete information, and/or with non-zero sum. Whenever score is relevant for success, SAI can contribute with parametric modeling of winrate as a function of targeted score, with branching training techniques, with real-time score estimates, with a parametric family of agents which allow real-time tuning the playing style to the opponent and to the game circumstances. For instance Puerto Rico~\cite{DRKM2017} could benefit from the general-purpose aspects of SAI. Finally, DRL is expected to extend to other domains outside of games, spanning from robotics to complex behaviors in 3D environments and character animation: as science progresses by contamination, we would not be surprised if SAI could be part of this conversation, too.

\section{Conclusion}

In conclusion, we have demonstrated the achievements of our framework in the \nine\ setting. Expanding to the \nineteen\ setting and experimenting with other games will allow to fully exploit the potential of SAI, both in the context of domain-independent Deep Reinforcement Learning, and as a means to progress towards the perfect Go game.  

\paragraph{Acknowledgements.} We thank Hayashi Kozo 6P and Oh Chimin 7D for accepting the challenge of playing with SAI; Alessandro Martinelli 2D, Davide Minieri 3D and Alessandro Pace 4D for support in the validation of suboptimal moves, and Michele Piccinno and Francesco Potortì\ for technical support.

\newpage

\bibliographystyle{plain}

\clearpage

\appendix

\def\wid{.25\textwidth}

\section{Match between SAI and Hayashi Kozo 6P}\label{s:games_hayashi}

The SGF files of these games are available as supplementary data attached to this arXiv version of the paper.

\begin{figure}[ht]
	\caption{Game 1. SAI (Black) vs Hayashi Kozo 6P (White). Black wins by 1.5 points.}\label{f:games_hayashi_1}
	\begin{center}
		\includegraphics[width=\wid]{\dirfig/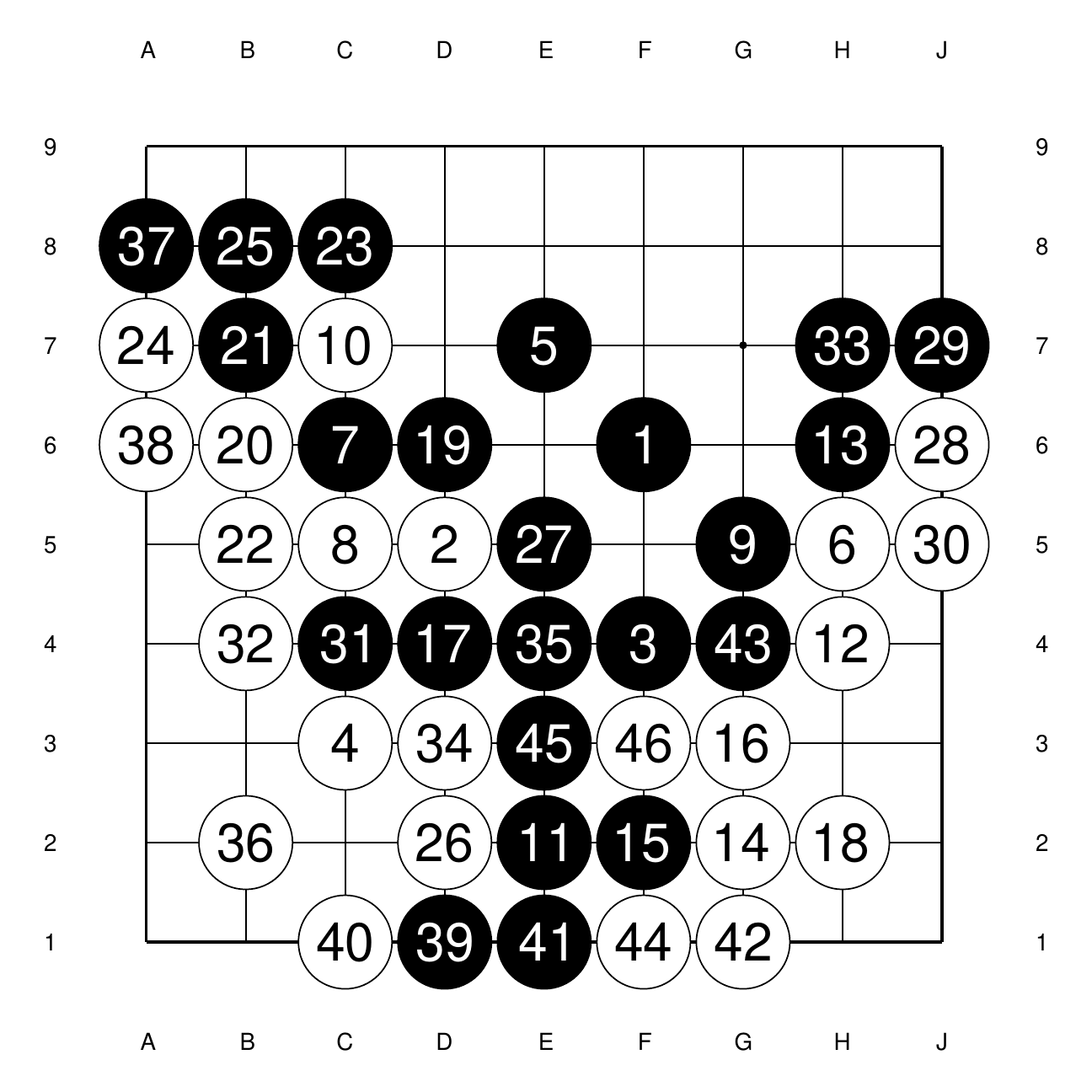}
	\end{center}
\end{figure}

\begin{figure}[ht]
	\caption{Game 2:  Hayashi Kozo 6P (Black) vs SAI (White). White wins by 6.5 points.}\label{f:games_hayashi_2}
	\begin{center}
		\includegraphics[width=\wid]{\dirfig/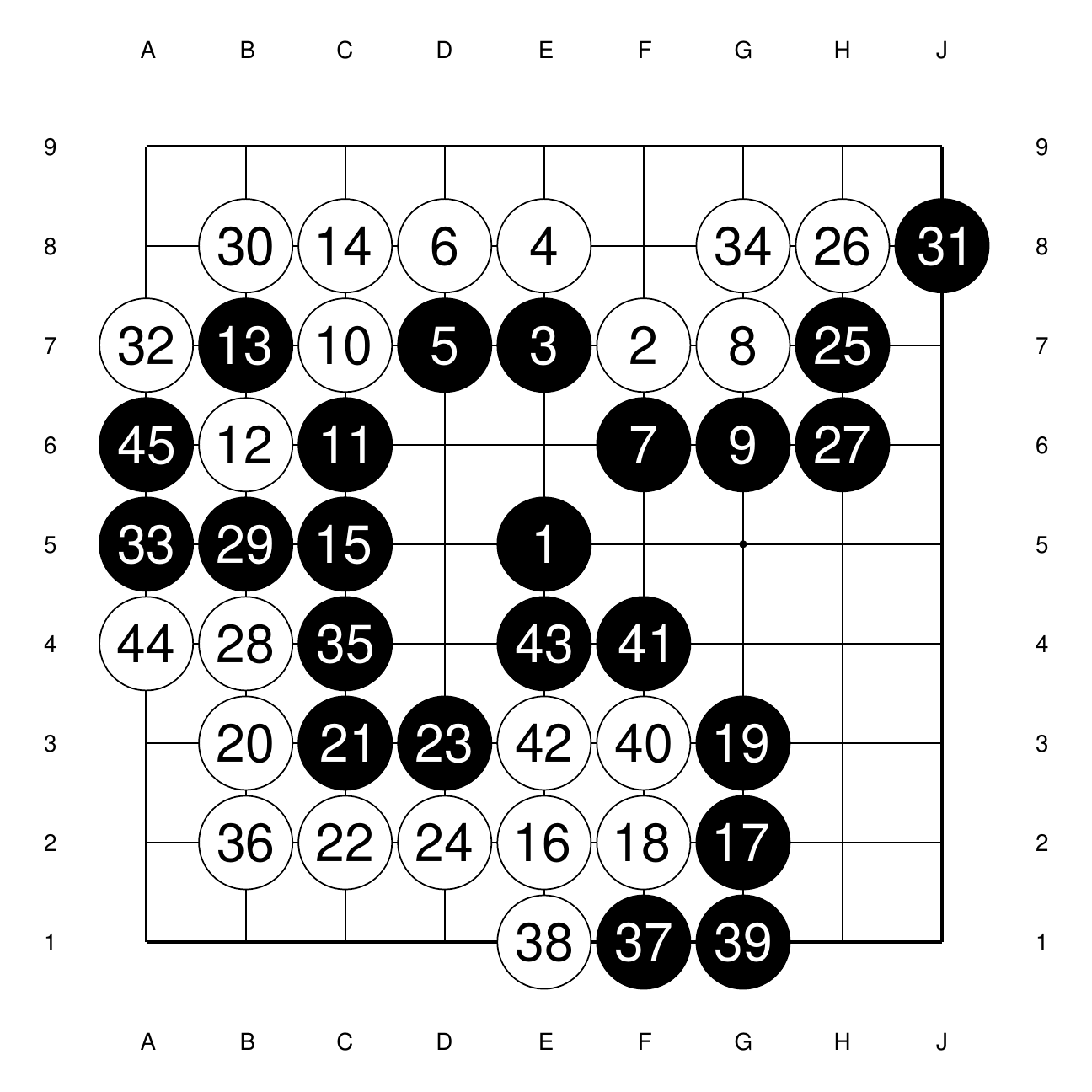} \\
		 \footnotesize 46 at 13
	\end{center}
\end{figure}

\begin{figure}[ht]
	\caption{Game 3: SAI (Black) vs Hayashi Kozo 6P (White). Black wins by resignation.}\label{f:games_hayashi_3}
	\begin{center}
		\includegraphics[width=\wid]{\dirfig/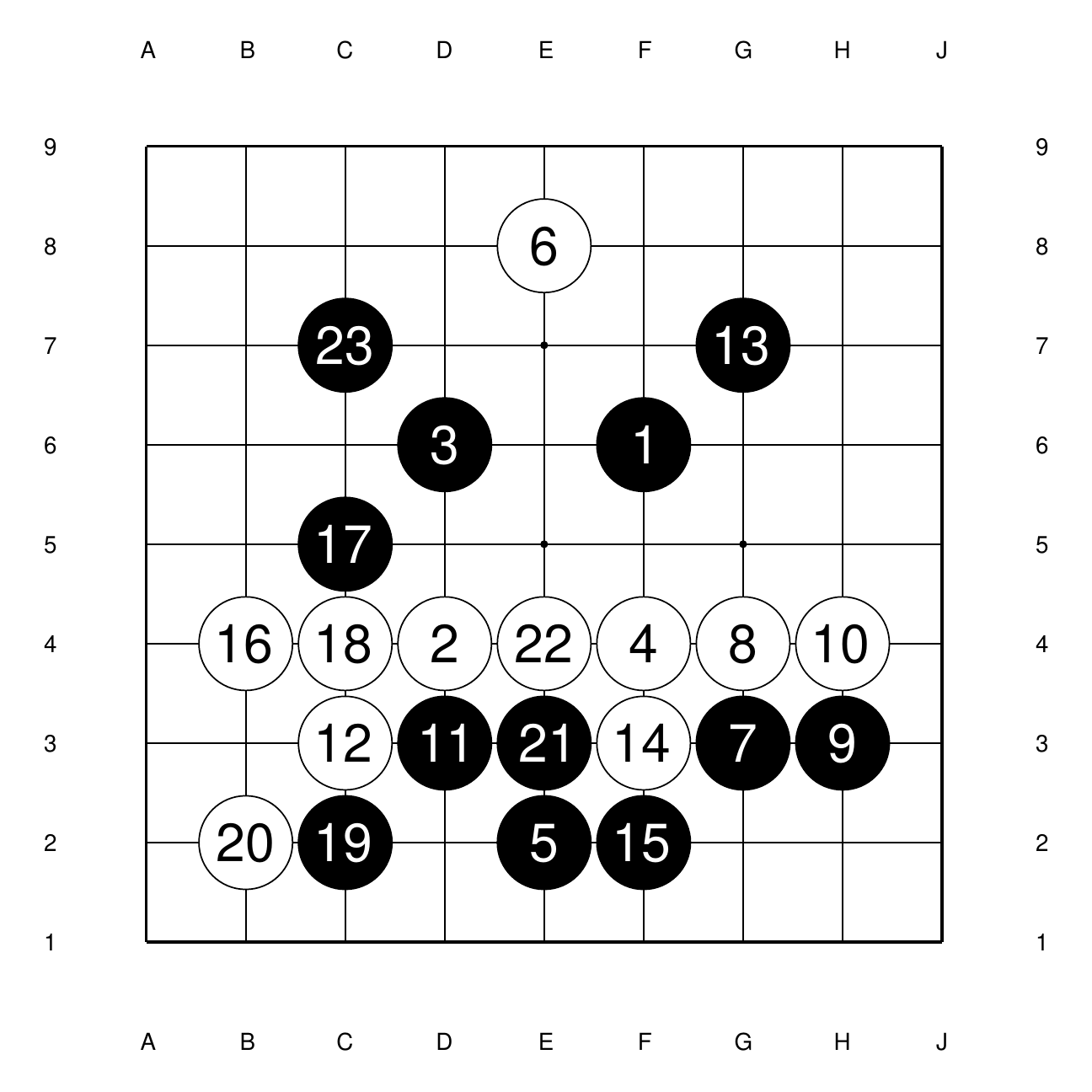}
	\end{center}
\end{figure}

\clearpage
\section{Match between SAI and Oh Chimin 7D}\label{s:games_OC}

The SGF files of these games are available as supplementary data attached to this arXiv version of the paper.

\begin{figure}[ht]
	\caption{Game 1. Oh Chimin 7D (Black) vs SAI (White). Komi 7.5. White wins by 2.5 points.}\label{f:games_OC_1}
	\begin{center}
		\includegraphics[width=\wid]{\dirfig/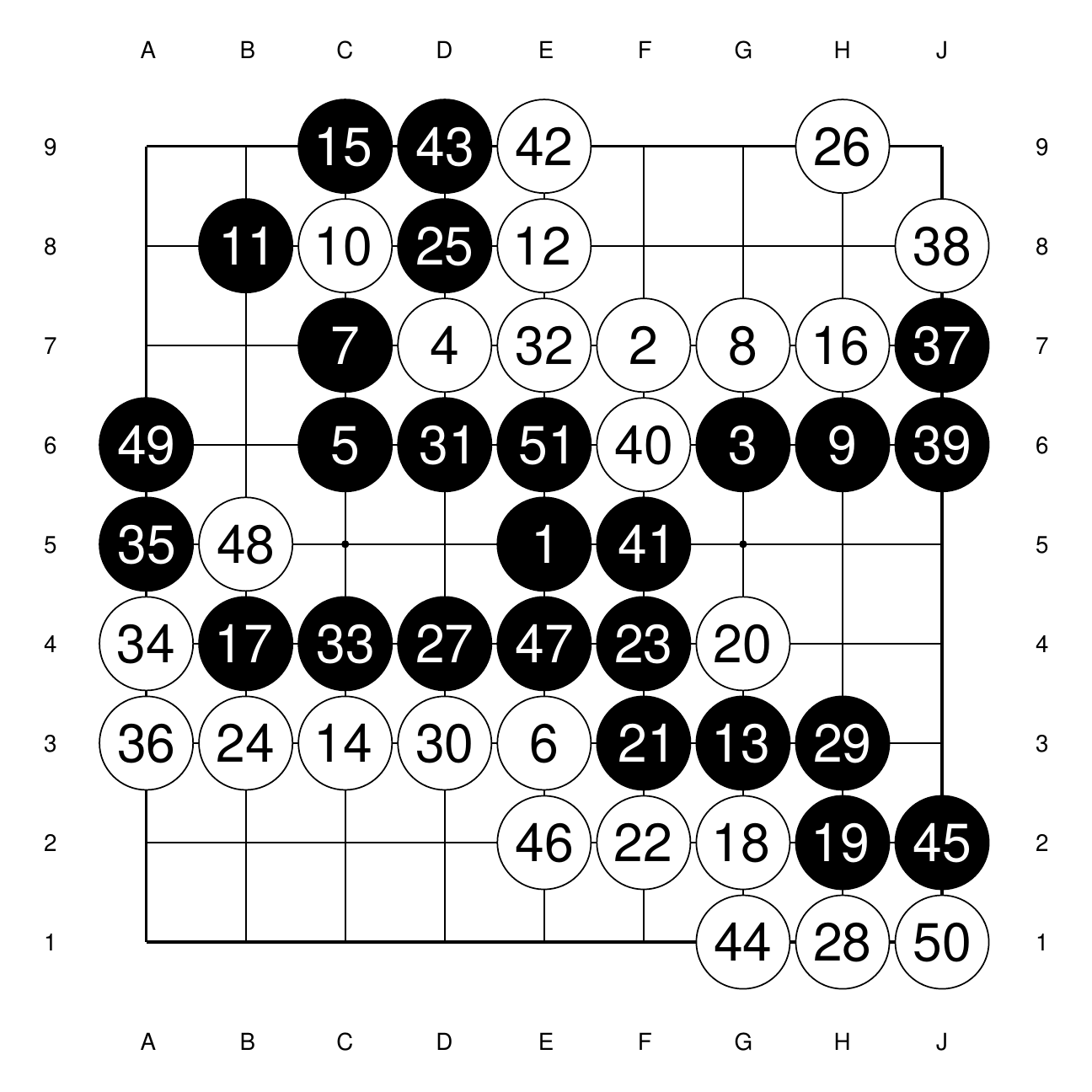}
	\end{center}
\end{figure}

\begin{figure}[ht]
	\caption{Game 2. Oh Chimin 7D (Black) vs SAI (White). Komi 5.5. White wins by resignation.}\label{f:games_OC_2}
	\begin{center}
		\includegraphics[width=\wid]{\dirfig/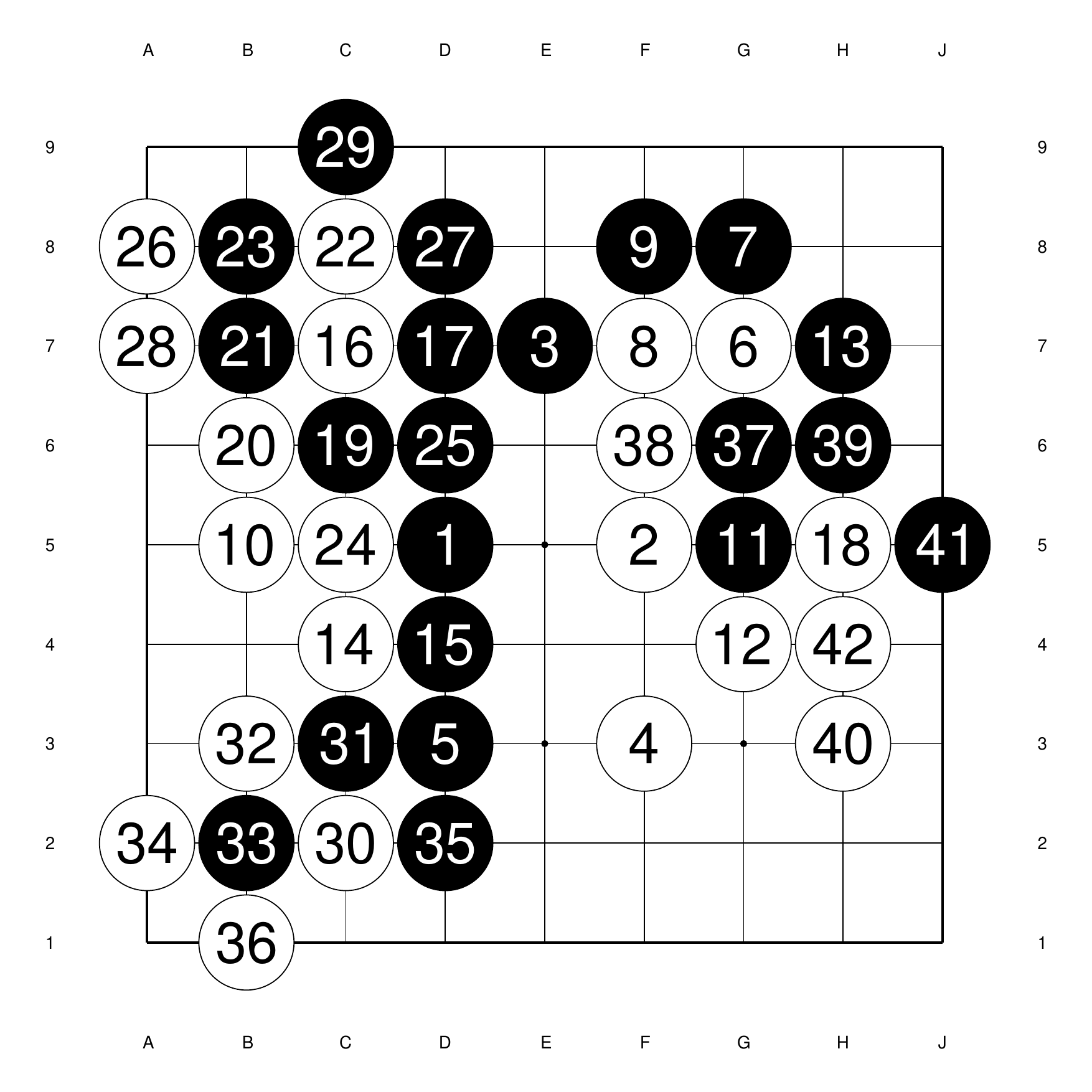}
	\end{center}
\end{figure}

\begin{figure}[ht]
	\caption{Game 3. Oh Chimin 7D (Black) vs SAI (White). Komi 3.5.  White wins by resignation.}\label{f:games_OC_3}
	\begin{center}
		\includegraphics[width=\wid]{\dirfig/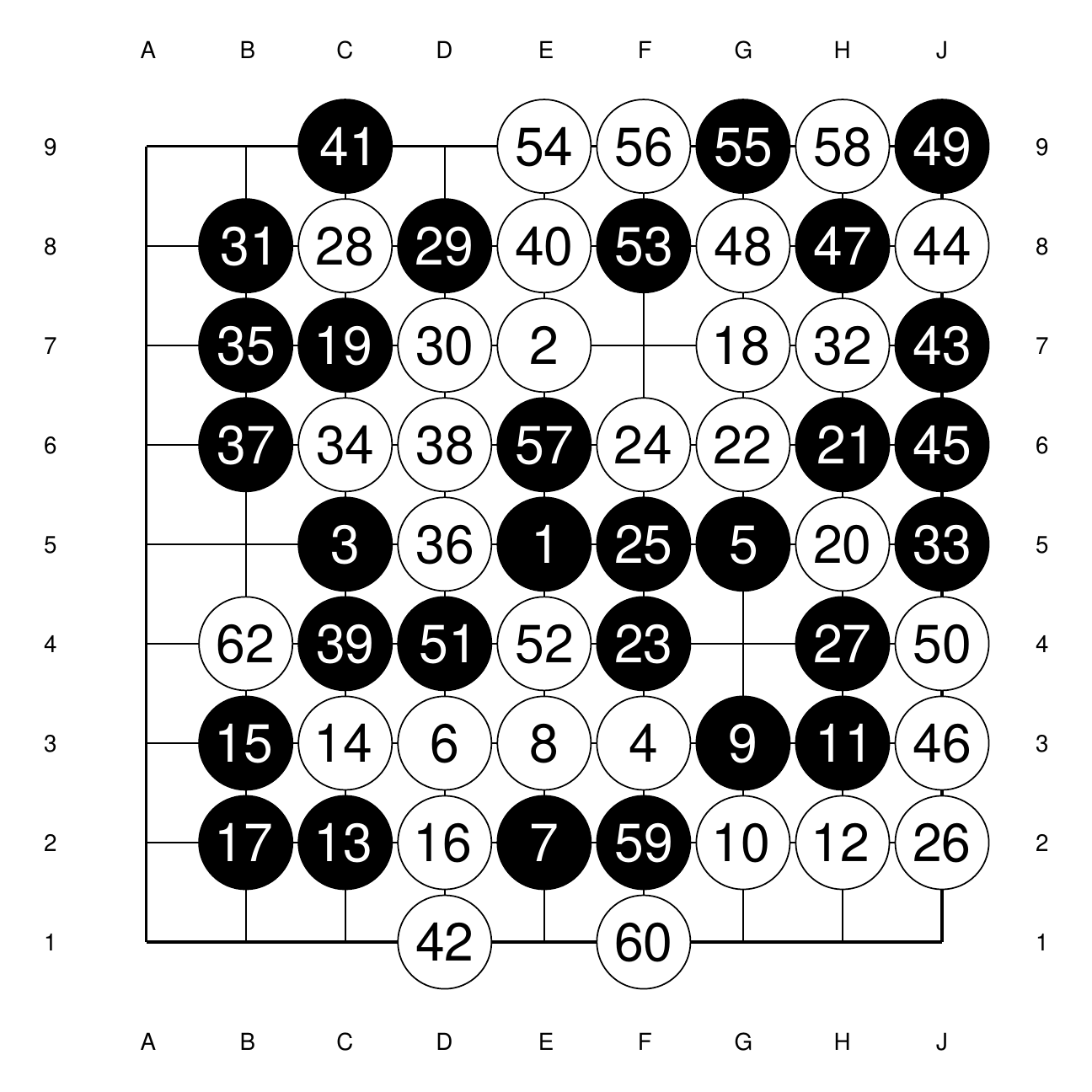}\\
		\footnotesize 61 at 55
	\end{center}
\end{figure}

\begin{figure}[ht]
	\caption{Game 4. Oh Chimin 7D (Black) vs SAI (White). Komi 1.5. Black wins by resignation.}\label{f:games_OC_4}
	\begin{center}
		\includegraphics[width=\wid]{\dirfig/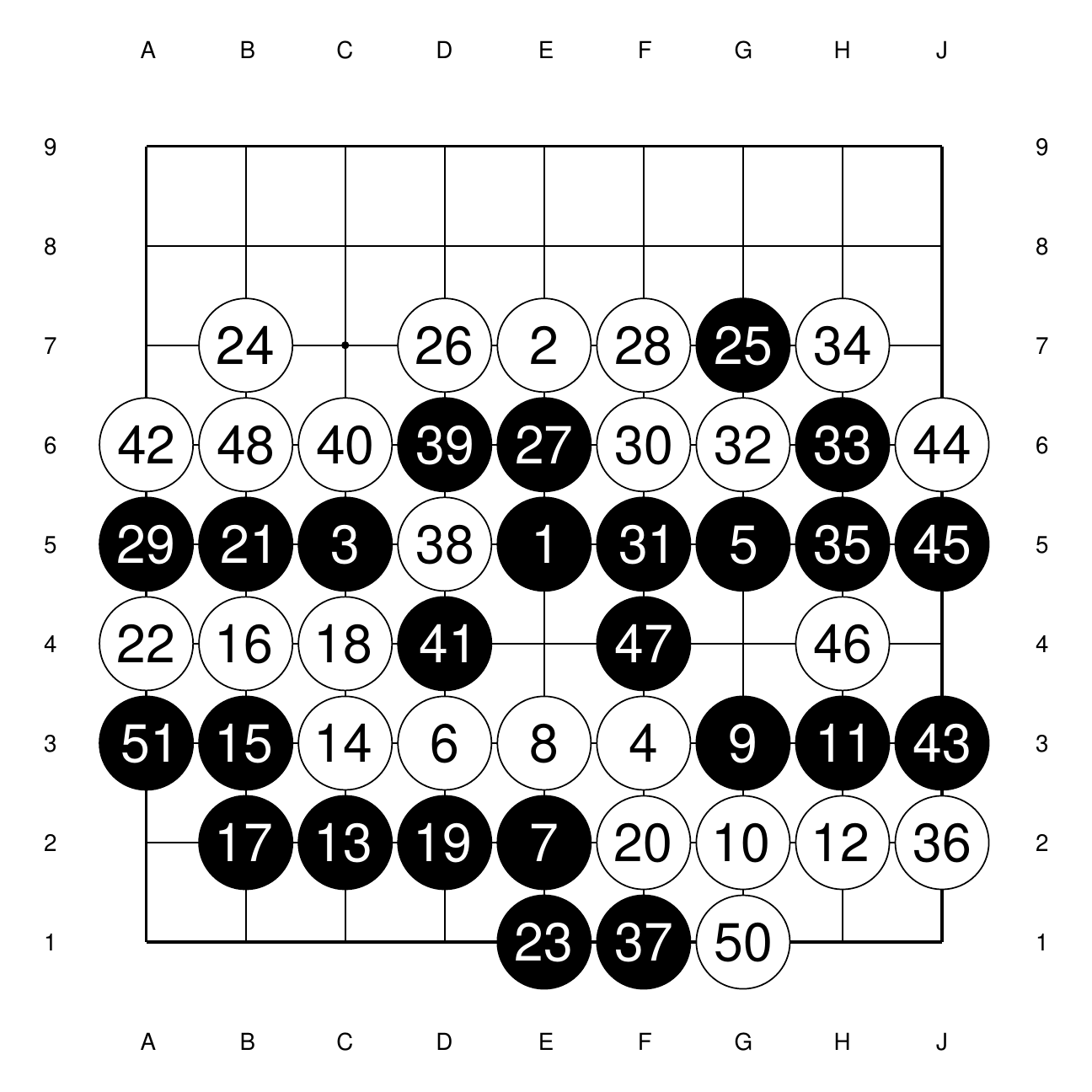}\\
		\footnotesize 49 at 38
	\end{center}
\end{figure}

\begin{figure}[ht]
	\caption{Game 5. SAI (Black) vs Oh Chimin 7D (white). Komi 13.5. Black wins by 1.5 points.}\label{f:games_OC_5}
	\begin{center}
		\includegraphics[width=\wid]{\dirfig/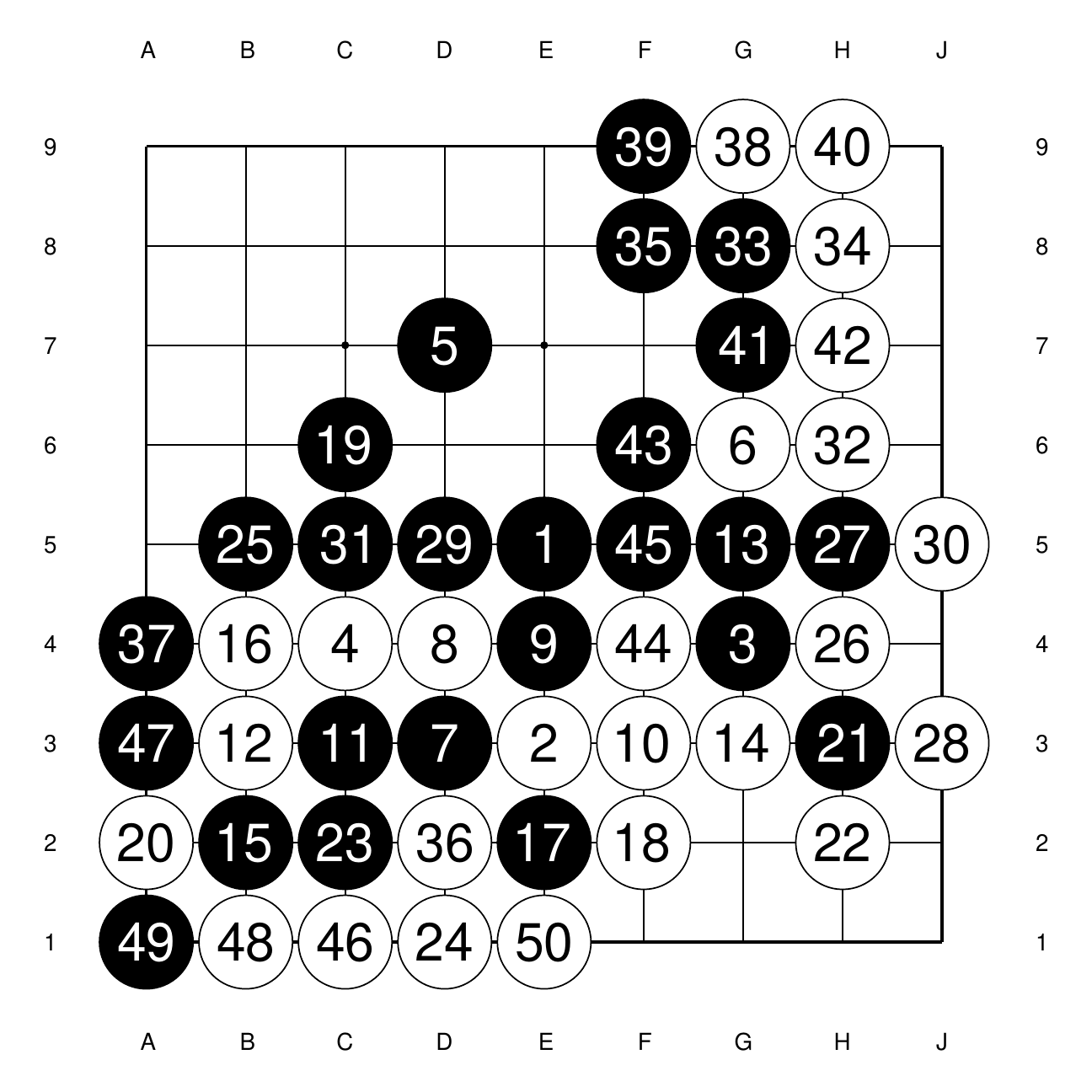}\\
		\footnotesize 51 at 20
	\end{center}
\end{figure}

\end{document}